\documentclass[11pt]{article}

\usepackage[final]{acl}

\usepackage{times}
\usepackage{latexsym}
\usepackage{tcolorbox}
\usepackage{booktabs}
\usepackage{amsmath}
\usepackage{multicol}
\usepackage{makecell}
\usepackage{multirow}
\usepackage{amssymb}
\usepackage{microtype}  
\usepackage{float}
\usepackage{siunitx}

\usepackage[T1]{fontenc}

\usepackage[utf8]{inputenc}

\usepackage{microtype}
\usepackage{comment}
\usepackage{inconsolata}
\usepackage{graphicx}
\usepackage{booktabs}
\usepackage{amsmath}
\usepackage{multicol}
\usepackage{tcolorbox}
\tcbuselibrary{skins, breakable}
\newcommand{\system}{UTP-Bench}

\title{\system: Uncertainty-aware Travel Planning Benchmark}

\author{ Etcharla Revanth Rao$^{1\dagger}$, Priyanshu Karmakar$^{1\dagger}$, Shubhojit Mallick$^2$, \\\textbf{Manish Gupta$^2$, Shreya Ghosh$^1$, Abhik Jana$^1$} \\
  $^1$IIT Bhubaneswar, India\quad $^2$Microsoft, India\\
\texttt{{a24cs08008,24CS06010,abhikjana,shreya}@iitbbs.ac.in}\\
  \texttt{{shubhojit.mallick,gmanish}@microsoft.com}
}

\begin{document}
\maketitle
\renewcommand{\thefootnote}{$\dagger$}
\footnotetext{The first two authors made equal contributions.}
\renewcommand{\thefootnote}{\arabic{footnote}}
\begin{abstract}

Large Language Models (LLMs) have recently demonstrated strong capabilities in automated travel itinerary generation. However, real-world travel planning is inherently \textit{uncertain}: transportation delays, crowd fluctuations, and unexpected stochastic delays frequently invalidate otherwise feasible schedules. Existing benchmarks like TravelPlanner and TripCraft assume deterministic environments, evaluating only static constraint satisfaction and ignoring whether generated plans remain robust when such uncertainties arise.

To address this limitation, we introduce \system{}\footnote{Code Base: \\\url{https://github.com/ETCHARLAREVANTHRAO/UTP-Bench}},
a large-scale benchmark for uncertainty-aware travel planning. The dataset integrates real-world travel data spanning 504 cities of India, including attractions, restaurants, accommodations, and multi-modal transportation networks. To model realistic disruptions, \system{} incorporates empirical delay distributions and crowd-density patterns collected from major cities, enabling evaluation of travel plans under stochastic conditions.

We further propose three evaluation metrics, namely \textit{Buffer Adequacy Score (BAS)}, \textit{Crowd-Aware Timing Score (CATS)}, and \textit{Transport Delay Absorption Score (TDAS)}, which quantify the ability of generated itineraries to maintain robustness against transit delays and crowd variability. Experiments with state-of-the-art LLMs like GPT-5, Qwen3, Mistral and Phi-4 reveal substantial gaps between model-generated and human-authored plans, particularly in temporal buffering, delay-aware transportation scheduling, and crowd-sensitive planning.

\end{abstract}

\section{Introduction}

Large Language Models (LLMs) have recently shown strong potential for automated travel itinerary generation, enabling personalized travel plans that satisfy user preferences, budget constraints, and temporal requirements. Recent works, including \textit{TravelPlanner}~\cite{xie2024travelplanner}, \textit{TripCraft}~\cite{chaudhuri2025tripcraft}, and \textit{Retail}~\cite{deng2025retail}, demonstrate that LLMs can generate structured multi-day itineraries while reasoning over diverse travel constraints such as attraction preferences, transportation connectivity, and scheduling requirements.

\begin{figure*}[]
\centering
\includegraphics[width=1.0\textwidth]{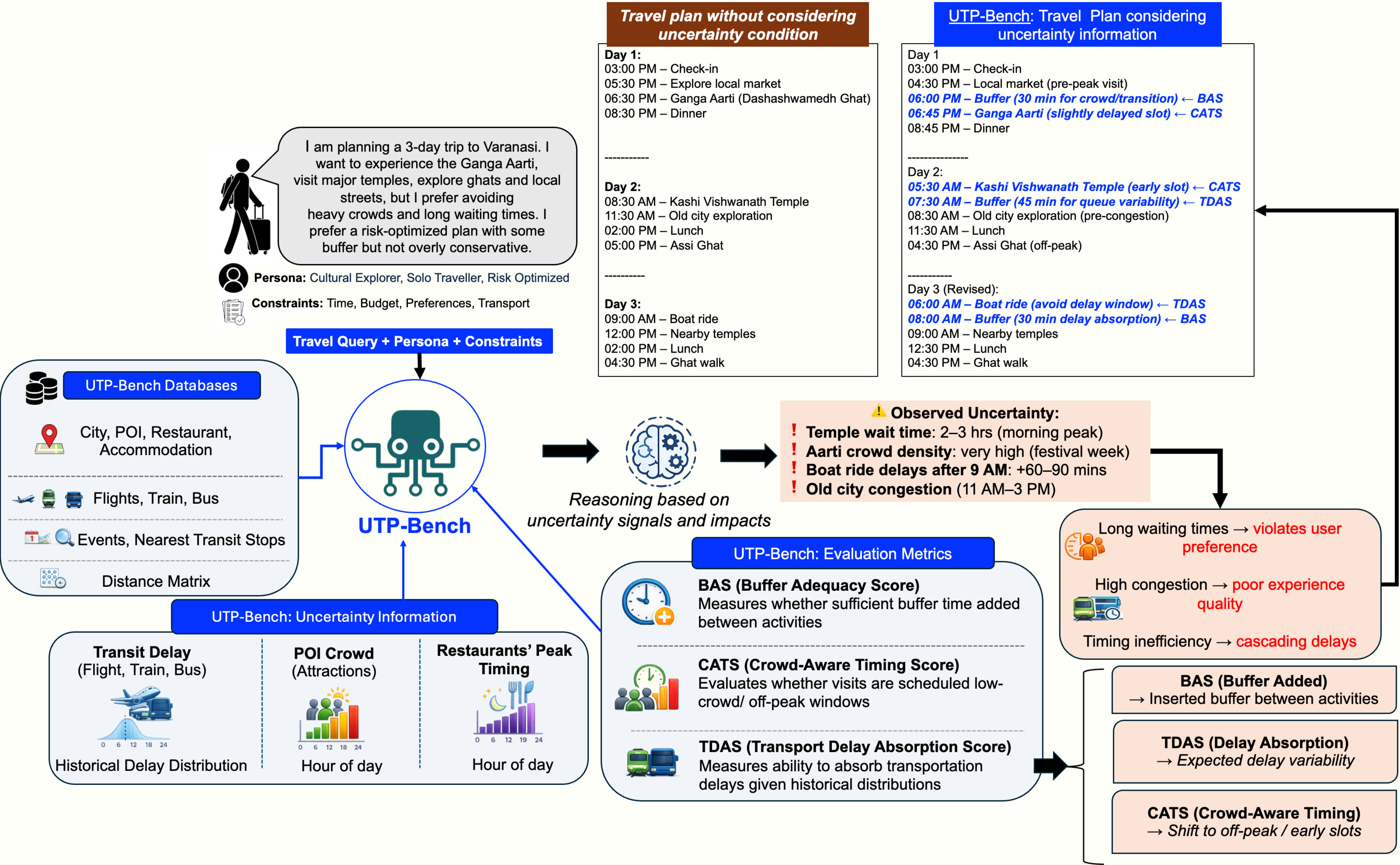}
\caption{\system: Bridging the gap between rigid travel plans and real-world uncertainty}
\label{fig:\system}
\end{figure*}

Despite recent advances, existing benchmarks remain limited in modeling real-world travel planning. First, they largely assume \textit{deterministic} environments with fixed transportation schedules, attraction durations, and transit conditions. In reality, travel planning is inherently uncertain due to delays, congestion, and unexpected disruptions, making otherwise feasible itineraries impractical. Consequently, evaluating plans solely through static constraint satisfaction fails to capture robustness under realistic conditions. Second, current benchmarks fail to represent the variability of real-world travel. Datasets such as \textit{TravelPlanner}, \textit{TripCraft}, and \textit{ChinaTravel}~\cite{shao2024chinatravel} rely on simplified assumptions regarding transit reliability and scheduling, while overlooking factors such as multimodal dependencies and dynamic crowd conditions that critically influence itinerary feasibility.

To address these limitations, we introduce \textbf{\system}, a benchmark for \textit{uncertainty-aware travel planning}. \system{} comprises 1000 real-world travel queries spanning 3-, 5-, and 7-day spatio-temporal fine-grained itineraries which is paired with a human-annotated gold-standard itinerary. Unlike prior datasets, \system{} incorporates empirical delay statistics and crowd-density patterns 
enabling evaluation of travel itineraries under realistic uncertainty.

Evaluating travel itineraries under uncertainty requires moving beyond binary feasibility checks. Existing approaches determine whether constraints are satisfied, but fail to assess how robust an itinerary remains under uncertainties. But what truly makes a travel plan reliable? \textit{Does the itinerary allocate sufficient transition time to absorb unexpected delays and variability in attraction visits? Can transportation schedules tolerate realistic disruptions without causing cascading failures across subsequent activities? Does the plan avoid highly congested periods, or does it schedule visits during peak crowd hours that may significantly affect the overall experience?} Most importantly, does the itinerary adapt its pacing according to different traveller risk preferences, balancing exploration with reliability?  To systematically assess these aspects, we introduce three uncertainty-aware evaluation metrics: \textit{Buffer Adequacy Score (BAS)}, which measures whether activity transitions include sufficient temporal buffers; \textit{Transport Delay Absorption Score (TDAS)}, which evaluates the resilience of transportation schedules under empirical delay distributions; and \textit{Crowd-Aware Timing Score (CATS)}, which quantifies the extent to which attraction visits avoid peak crowd periods.
Furthermore, \system{} models diverse traveler behaviors through \textit{risk-aware travel profiles}, including \textit{Risk-Tolerant}, \textit{Risk-Optimized}, and \textit{Risk-Averse} travelers. These profiles capture different levels of tolerance to scheduling uncertainty and influence the amount of buffer time and itinerary density that are considered acceptable. Our \textbf{contributions} are three-fold: 

\noindent\textbf{1. Uncertainty-aware travel planning benchmark.} We introduce \system, a large-scale travel planning dataset integrating real-world attractions, restaurants, accommodations, and multimodal transportation data. 

 \noindent\textbf{2. Modeling real-world travel uncertainty and traveler profiles.} \system{} incorporates empirical transit delay statistics and crowd-density patterns to enable systematic evaluation of travel plans with uncertainties.  The benchmark also introduces risk-aware traveler profiles that model varying tolerance to scheduling uncertainty, supporting more realistic itinerary generation and evaluation. 
 
 \noindent \textbf{3. Novel evaluation metrics for robust travel planning.} We propose BAS, TDAS, and CATS to quantify the robustness of generated itineraries with respect to temporal buffers, crowd dynamics, and transportation delays.

\section{Related Work}

\noindent\textbf{Planning with Large Language Models.} Large language models (LLMs) have shown strong capabilities in planning tasks, including scheduling, commonsense reasoning, and heuristic-guided decision-making \cite{valmeekam2023planning, pallagani2023understanding, prasad2024adapt, lee2025evolving}. Prior work has improved planning performance through grounded few-shot planning \cite{song2023llmplanner} and integration with classical search methods such as Monte Carlo Tree Search \cite{coulom2006efficient, swiechowski2023monte, zhao2024large}. However, LLMs still struggle to generate reliable plans in open-ended domains due to challenges in modeling subgoal dependencies, handling cascading constraint violations, and reasoning under uncertainty \cite{valmeekam2022planbench, kambhampati2023role}. \\
\noindent\textbf{LLMs in Travel Planning.} Travel planning requires satisfying spatial, temporal, budgetary, and preference-driven constraints over extended horizons \cite{xi2025rise, jonnala2025exploring}. Prior benchmarks, including TravelPlanner \cite{xie2024travelplanner}, TravelPlanner+ \cite{singh2024personal}, and TripCraft \cite{chaudhuri2025tripcraft}, advanced constraint-aware and persona-driven itinerary generation using real-world data. However, these approaches assume deterministic conditions, overlooking uncertainties such as transit delays, variable visit durations, and crowd dynamics. While TripTide \cite{karmakar2025triptidebenchmarkadaptivetravel} introduces disruptions, its scenarios remain synthetic. In contrast, \system{} models travel planning as a stochastic problem grounded in empirical uncertainty data.
\\
\noindent\textbf{Evaluation of LLM-Generated Travel Plans.} 
Early travel-planning benchmarks primarily relied on binary constraint satisfaction~\cite{xie2024travelplanner}, later extended with personalization and itinerary-quality metrics~\cite{chen2024travelagent,singh2024personal}. TripCraft~\cite{chaudhuri2025tripcraft} further introduced fine-grained spatial, temporal, and persona-based evaluation. However, existing frameworks remain deterministic and overlook real-world uncertainty such as transit delays and crowd congestion. \system{} addresses this gap through three uncertainty-aware metrics (\textit{BAS}, \textit{TDAS}, \textit{CATS}) that evaluate itinerary robustness under empirical uncertainty distributions. To our knowledge, \system{} is the first benchmark to explicitly model stochastic robustness in LLM-based travel planning.
\section{\system{} Dataset Curation}
\textbf{\system{}} is a benchmark for evaluating LLMs in uncertainty-aware, constraint-rich travel itinerary planning. It assesses whether models can generate itineraries that are feasible, preference-aligned, and robust to real-world uncertainties, including transit delays, crowd fluctuations, and dynamic scheduling conditions. 
The benchmark comprises \textbf{1,000 travel queries} spanning 3-, 5-, and 7-day itineraries across diverse planning scenarios of varying complexity. Each query is paired with a human-annotated gold-standard itinerary, alongside model-generated itineraries produced under uncertainty-aware prompting settings.

\begin{table}[!h]
\centering
\scriptsize

\begin{tabular}{l r}
\toprule
\textbf{Database} & \textbf{Data Entries (\#)} \\
\midrule
City Set & 504 \\
States \& Union Territories & 33 \\
Flights$^{\dagger}$ & 29,346 \\
Trains$^{\dagger}$ & 68,068 \\
Buses$^{\dagger}$ & 34,925 \\
Restaurants$^{\dagger}$ & 3,433 \\
Attractions$^{\dagger}$ & 2,990 \\
Accommodations & 3,670 \\
Events & 783 \\
Nearest Transit Stop & 10,025 \\
Distance Matrix & 253,513 \\
\bottomrule
\end{tabular}
\caption{Dataset statistics in \system{}. Components marked with $\dagger$ include uncertainty-aware signals}
\label{tab:db_entries1}
\end{table}

\subsection{Uncertainty-Aware Data Curation}
The benchmark is constructed from large-scale travel data covering 504 cities across 33 Indian states and union territories (Table~\ref{tab:db_entries1}). Data is collected from real-world sources through web scraping and open-source resources such as OpenStreetMap (Appendix~\ref{app:data-sourcing}). The database includes city-level information on transportation, attractions, restaurants, accommodations, and related metadata, with incomplete records removed or normalized for consistency.

To model realistic travel uncertainty, we augment the dataset with empirical transit-delay statistics and crowd-density patterns. Specifically, we collect historical delay data for flights, trains, buses, and cabs, along with crowd-density information for attractions and restaurants across 20 major Indian cities (Appendix~\ref{app:data-sourcing}). These uncertainty signals capture real-world variability that can undermine otherwise feasible itineraries, such as cascading delays or congestion during peak hours. By incorporating such factors, \system{} enables evaluation of itinerary robustness beyond constraint satisfaction.

\subsection{Constraints and Persona Data}
\label{sec:constraints}

\system{} builds upon the constraint and persona framework introduced in TripCraft~\cite{chaudhuri2025tripcraft}, while extending it with an explicit uncertainty-aware layer. The benchmark includes commonsense constraints, hard constraints, persona information, and newly introduced \textit{uncertainty constraints that are absent from prior benchmarks}.

\noindent\textbf{Uncertainty Constraints.}

These constraints arise from two major sources. First, \textbf{transit delays}: itineraries must allocate sufficient buffer to absorb empirically observed delays across flights, trains, buses, and cabs. Second, \textbf{crowd-induced variability}: visit durations at attractions and restaurants depend on the time of day and the expected crowd intensity, so plans that schedule activities during highly congested periods without appropriate adjustments are penalized. These uncertainty constraints form the basis of our uncertainty-aware evaluation metrics. 

\noindent\textbf{Persona Information.}
Each query in \system{} is paired with a persona profile that influences itinerary generation along four dimensions: traveler type, purpose of travel, spending preference, and location preference. Traveler type distinguishes, for instance, between laid-back and adventure-oriented travelers; purpose of travel captures motivations such as relaxation, adventure, or cultural exploration; spending preference shapes cost-sensitive decisions; and location preference reflects favored environments such as beaches, mountains, heritage cities, wildlife destinations, or religious sites. 
In addition, \system{} augments each query with a \textbf{risk profile}, which governs how much uncertainty a traveler is willing to tolerate. A \textit{Risk-Tolerant} traveler accepts tighter schedules and smaller buffers, a \textit{Risk-Averse} traveler prefers conservative timing and larger recovery margins, and a \textit{Risk-Optimized} traveler lies between these two extremes. This allows the benchmark to evaluate not only whether a plan is feasible, but also whether it is appropriately calibrated to the traveler’s uncertainty tolerance (See Table~\ref{tab:constraints}).

\noindent\textbf{Commonsense and Hard Constraints.}
These constraints ensure basic itinerary realism and coherence. For example, events cannot repeat across days, meals must maintain appropriate temporal gaps, and PoIs must follow a valid chronological order. Additionally, except for the final departure day, each itinerary day begins and ends at the designated accommodation with feasible transitions across activities. \system{} further enforces hard constraints on travel feasibility and diversity (e.g., budget, room rules, and accommodation type)~\cite{xie2024travelplanner}. To encourage activity diversity, attractions are mapped to 20 semantic categories spanning different travel interests.

\begin{table}[!h]
\centering
\scriptsize

\tabcolsep3pt
\begin{tabular}{l p{0.64\columnwidth}}
\toprule
\textbf{Constraint} \\
\midrule
\multicolumn{2}{l}{\textbf{Uncertainty Constraints }} \\
Transit Delay Buffer & Buffers must account for empirical delay distributions across flights, trains, buses, and cabs. \\
Crowd-Induced Delay & Visit and wait times must reflect peak-hour crowd density patterns. \\
Activity Buffer Gap & Gaps between activities must absorb real-world transit and crowd disruptions. \\
\midrule
\multicolumn{2}{l}{\textbf{Risk Profiles}} \\
Risk-Tolerant & Tighter scheduling with limited additional buffer. \\
Risk-Optimized & Balanced schedule with moderate buffer allocation. \\
Risk-Averse & Conservative planning with larger safety margins. \\
\bottomrule
\end{tabular}
\caption{Constraint and persona details in \system{}. All hard and commonsense constraints are retained from TripCraft, TravelPlanner (See Table~\ref{tab:full_constraints}, Appendix~\ref{app:prompts}).}
\label{tab:constraints}
\end{table}

\subsection{Query Construction}

Travel queries are created by sampling a structured set of planning inputs, including departure city, destination, travel dates, budget, and persona profile. Trip duration determines the geographic scope of the itinerary: 3-day plans focus on a single destination city, while 5-day and 7-day plans cover one state with visits to two and three cities, respectively. Each query is further associated with hard constraints, persona attributes, and a risk profile. These components are combined using GPT-5 in a few-shot setting to produce naturalistic travel-planning queries that are both diverse and structurally controlled. To capture different levels of planning difficulty, queries are categorized as Easy, Medium, or Hard based on the density of available PoIs, complexity of transit connectivity, and the degree of scheduling constraints required by the itinerary using GPT5. 
\begin{figure*}[t]
    \centering
    \includegraphics[width=0.85\linewidth]{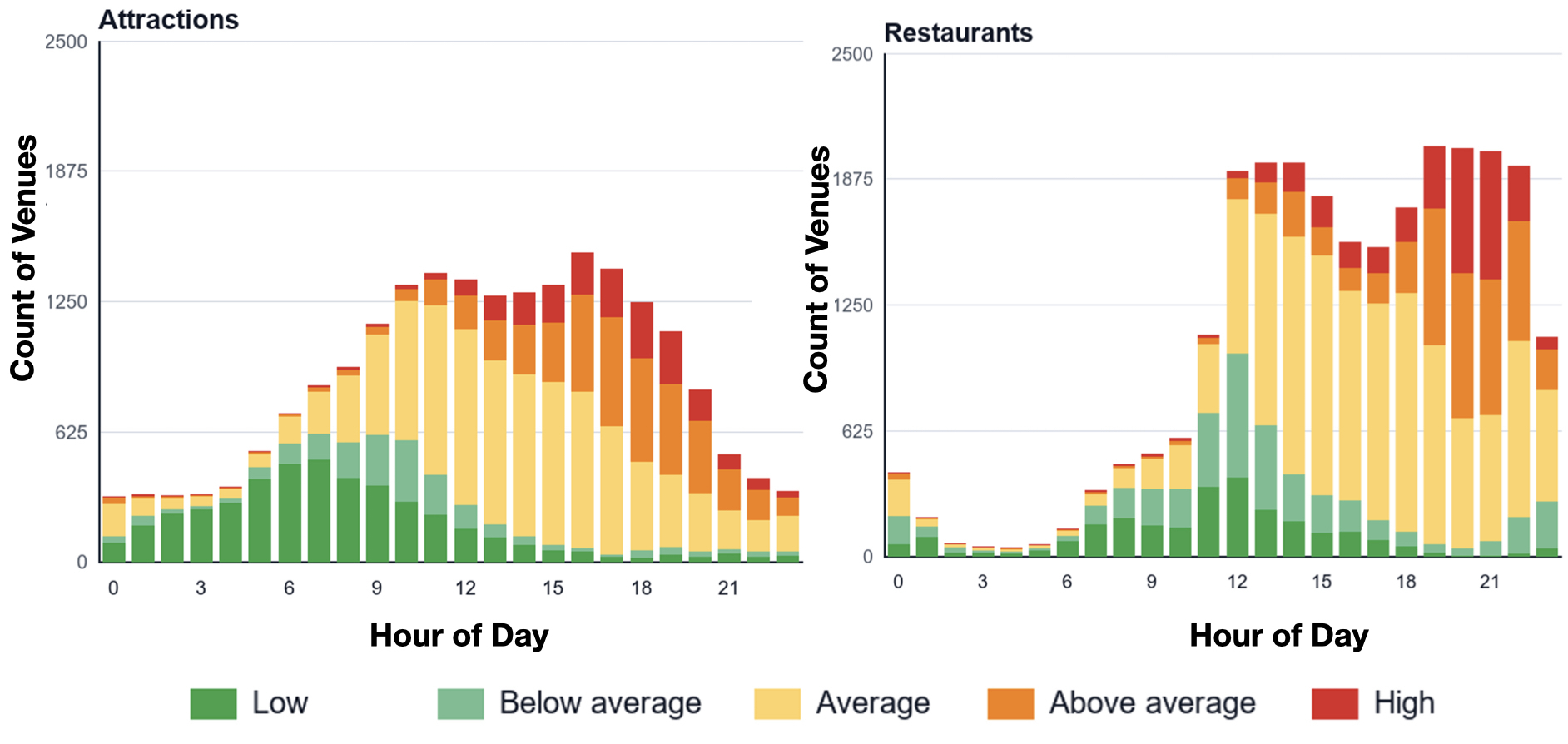}
    \caption{Hourly crowd-level distributions across attractions and restaurants. Each bar represents the number of venues within different crowd categories at a given hour.}
    \label{fig:crowd_levels}
\end{figure*}
\subsection{Annotation and Refinement}

Each query is paired with a gold-standard itinerary created by 14 trained human annotators through multiple rounds of refinement. Our annotation process is not purely manual: annotators are supported by \textit{purpose-built helper scripts} that surface relevant contextual information during planning, including historical transit-delay statistics, crowd-density patterns, and expected attraction-visit durations, drawn directly from the database. This automated script-assisted workflow allows annotators to produce itineraries that are not only logically valid and preference-aligned, but also genuinely uncertainty-aware. In particular, annotators must reason about temporal feasibility, spatial coherence, crowd-sensitive scheduling, transit-delay absorption, and the risk profile associated with each query. Each annotated itinerary is accompanied by a written rationale that explains the key planning decisions, the constraint-satisfaction logic, and the uncertainty-buffer choices, thereby improving interpretability. See annotation guidelines in Appendix~\ref{app:annotator-details}, Table~\ref{tab:annotation_guidelines}.

\subsection{Quality Control}

To ensure the reliability of the benchmark, we adopt a multi-stage quality-control process. First, annotation proceeds through iterative refinement rounds, during which annotators revise itineraries based on feedback from domain experts. Second, all query-plan pairs undergo final manual review to verify feasibility, optimality, and consistency with the intended persona and risk profile. Third, we use automated evaluation scripts to check structural validity, including temporal consistency, entity correctness, and compliance with key itinerary constraints. This combination of human refinement, domain-expert review, and script-based verification ensures that the benchmark captures realistic travel-planning behavior while maintaining high annotation quality.
\subsection{Analysis of Uncertainty Distribution}

Transportation uncertainty in our framework is modeled using historical delay signals across rail, flight, and road travel. Train delays range from 40.8-46.7 mins on average, while delayed flights exhibit a mean delay of 40.59 mins and road travel shows the highest average delay (52.2 mins). Across all modes, mean delays consistently exceed median delays, indicating skewed distributions with occasional large disruptions. Figure~\ref{fig:crowd_levels} further illustrates temporal crowd dynamics, showing that attractions experience higher crowd density during daytime and afternoon hours, whereas restaurants exhibit stronger evening concentration. Together, these observations highlight the strong temporal dependence of transportation delays and crowd patterns, motivating uncertainty-aware itinerary generation.

\if{0}
\begin{figure*}[t]
    \centering
    \includegraphics[width=0.9\linewidth]{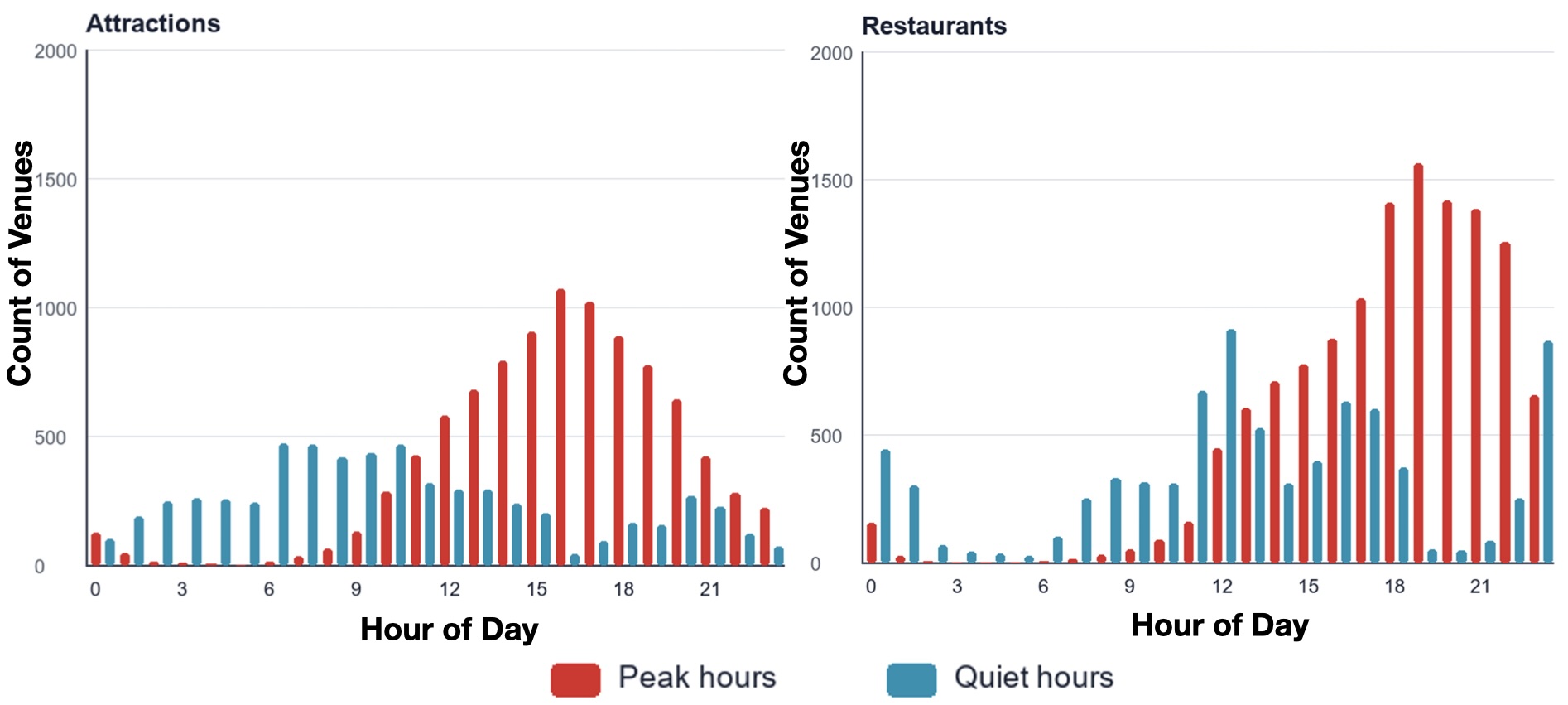}
    \caption{Peak-hour and quiet-hour distributions for attractions and restaurants. Each bar represents the number of venues having Peak or Quiet crowd at a given hour.}
    \label{fig:peak_quiet}
\end{figure*}
\fi

\section{Evaluation Metrics}
Plan feasibility is assessed using the hard, commonsense, and uncertainty constraints mentioned in Section~\ref{sec:constraints}. \system{} uses a comprehensive eight-metric evaluation suite that captures temporal, spatial, sequential, persona-specific, and uncertainty-aware nuances of a travel plan.

We adopt five metrics, namely \textit{Temporal Meal Score}, \textit{Temporal Attraction Score}, \textit{Spatial Score}, \textit{Persona Score}, and \textit{Ordering Score} from TripCraft \cite{chaudhuri2025tripcraft}, retaining their formulations and parameter derivation methodology. Further, we introduce three novel uncertainty-aware metrics, namely  \textit{Buffer Adequacy Score (BAS)}, \textit{Crowd-Aware Timing Score (CATS)}, and \textit{Transport Delay Absorption Score (TDAS)} to evaluate plan robustness under the stochastic disruptions characteristic of real-world travel.

\noindent\textbf{Buffer Adequacy Score (BAS):}
The Buffer Adequacy Score evaluates the \textit{temporal realism} of an itinerary by assessing whether the allocated buffer time at each Point of Interest (POI) is consistent with its expected visit duration, local crowd conditions, and the traveler’s risk tolerance.

For a POI with arrival time $t_a$ and departure time $t_d$, the actual buffer time $B_a$ is defined as:
\begin{equation}
  B_a = (t_d - t_a) - v
\end{equation}
where $v$ denotes the base visit duration of the POI.

To account for uncertainty, we define an expected buffer range $[a_c, b_c]$ that adapts to both crowd congestion and traveler preferences:
\begin{equation}
  \begin{split}
    a_c &= \mathit{low\_mult} \cdot v \cdot C_f \\
    b_c &= \mathit{high\_mult} \cdot v \cdot C_f
  \end{split}
\end{equation}
where $C_f$ represents the crowd congestion factor, and $\mathit{low\_mult}, \mathit{high\_mult}$ are scaling parameters determined by the traveler’s risk profile. We ensure that $high\_mul>low\_mul$ and $C_f>1$.

We define a deviation penalty $Q_i$ that quantifies the extent to which the allocated buffer deviates from the acceptable range. We use min rather than max because max would measure the distance to the farther boundary, which would over-penalize the itinerary.
\begin{equation}
\scriptsize
  Q_i =
  \begin{cases}
    0, & \text{if } B_a \in [a_c, b_c] \\[6pt]
    \dfrac{\min\big(|B_a - a_c|,\ |B_a - b_c|\big)}{b_c - a_c}, & \text{otherwise}
  \end{cases}
\end{equation}

Finally, for an itinerary containing $N$ POI transitions, the Buffer Adequacy Score is computed as:
\begin{equation}
  \mathrm{BAS} = 1 - \frac{1}{N} \sum_{i=1}^{N} min(1, Q_i)
\end{equation}

\noindent A BAS value closer to $1$ indicates that the itinerary allocates buffer times are well-aligned with expected uncertainty, while lower values reflect either overly tight or excessively conservative scheduling.

\noindent\textbf{Crowd-Aware Timing Score (CATS).}
The Crowd-Aware Timing Score evaluates how effectively POI visits are scheduled to avoid peak congestion and align with low-crowd periods. For each POI visit, we compute a \textit{crowd intensity} score based on the overlap of the visit duration with predefined crowd windows. This is obtained as a duration-weighted average of crowd levels across the visited time intervals.  Details in Appendix~\ref{app:cats}. 
This assigns higher scores to visits occurring during less crowded periods. To further refine this measure, we incorporate two adjustments: (i) a \textit{quiet-hour bonus}, which rewards visits overlapping with low-density time windows, and (ii) a \textit{peak-hour penalty}, which penalizes visits scheduled during high-density periods. These adjustments are applied proportionally to the fraction of time spent in quiet or peak intervals.

The final score for each POI is normalized to the range $[0,1]$, where higher values indicate better alignment with favorable crowd conditions. We convert this into a penalty term $score_i$ as detailed in Appendix~\ref{app:cats}

For an itinerary with $N$ POI visits, the overall Crowd-Aware Timing Score is computed as:
\begin{equation}
  \mathrm{CATS} = \frac{1}{N} \sum_{i=1}^{N} score_i
\end{equation}
A higher CATS value indicates that the itinerary consistently schedules visits during less crowded periods, reflecting better crowd-aware planning.

\noindent\textbf{Transport Delay Absorption Score (TDAS).}
The Transport Delay Absorption Score evaluates whether the buffer allocated for each transport segment is sufficient to absorb realistic delays without propagating disruptions to subsequent activities in the itinerary.

For a transport segment $j$, the planner implicitly allocates a buffer defined as the difference between the planned travel duration and the corresponding historical duration:
\begin{equation}
  B_j = \mathit{planned\_duration}_j - \mathit{hist\_duration}_j
\end{equation}

To assess adequacy, we estimate an \textit{expected delay} $E[D]$ for each segment based on historical data, which varies across transport modes (e.g., flights, trains, and road-based travel). This expected delay captures both the likelihood and magnitude of delays observed in practice.

Given $E[D]$, we define an acceptable buffer range that scales with the traveler’s risk tolerance. Specifically, lower and upper bounds are obtained by applying risk-dependent multipliers to $E[D]$, reflecting how conservative or aggressive the itinerary is with respect to delay absorption. We then compute a deviation penalty $Q_j$ for each segment, which is zero when the allocated buffer lies within the acceptable range, and increases proportionally as the buffer deviates from this range. This captures both under-allocation (fragile plans) and over-allocation (inefficient plans).

For an itinerary with $M$ transport segments, the overall TDAS is defined as:
\begin{equation}
  \mathrm{TDAS} = 1 - \frac{1}{M} \sum_{j=1}^{M} min(1,Q_j)
\end{equation}

A higher TDAS value indicates that the itinerary allocates buffers that are well-aligned with expected transport delays. (Details in Appendix~\ref{app:more-evaluation}) 

\section{Experimental Results and Analysis}
We adopt the direct sole planning strategy~\cite{xie2024travelplanner,singh2024personal}, modifying the prompt to include PoI lists, uncertainty parameters, and a refined one-shot example tailored to our constraints. 
Additionally, we incorporate 
natural language descriptions of the parameterised distributions modelling the metrics

including traveller risk profiles, congestion tiers, and transport delay statistics in the prompt (details in Appendix~\ref{app:prompts}). We evaluate GPT-5~\citep{singh2026openaigpt5card}, Qwen3~(14B)~\citep{yang2025qwen3technicalreport}, Mistral-7B-Instruct-v0.3\citep{jiang2023mistral7b} and Phi-4 mini-Instruct~\citep{microsoft2025phi4minitechnicalreportcompact} across 3-day, 5-day, and 7-day travel plans using the \system 
\footnote{Since we are not training or fine-tuning LLMs, we directly report results without train/validation splits.}. 
Results are summarized in Tables~\ref{tab:combined_metrics1}.

\begin{table*}[!h]
\centering
\scriptsize
\setlength{\tabcolsep}{3pt}

\resizebox{\textwidth}{!}{%
\begin{tabular}{l l r r r r r r r r r r r r r r r r r r r r r}
\toprule
\textbf{Model} & \textbf{Category} & \textbf{\makecell{Delivery\\Rate (\%)}} & \multicolumn{2}{c}{\textbf{CPR (\%)}} & \multicolumn{2}{c}{\textbf{HCPR (\%)}} & \textbf{\makecell{Final Pass\\Rate (\%)}} & \textbf{$T_{\text{meal}}$} & \textbf{$T_{\text{attr}}$} & \textbf{$S_{\text{spatial}}$} & \textbf{$P_{\text{persona}}$} & \textbf{$O_{\text{ord}}$} & \multicolumn{3}{c}{\textbf{BAS(\%)\,$\uparrow$} }& \multicolumn{3}{c}{\textbf{CATS(\%)\,$\uparrow$} }& \multicolumn{3}{c}{\textbf{TDAS(\%)\,$\uparrow$}} \\
\cmidrule(lr){4-5} \cmidrule(lr){6-7} \cmidrule(lr){14-16} \cmidrule(lr){17-19} \cmidrule(lr){20-22}
& & & \textbf{Micro} & \textbf{Macro} & \textbf{Micro} & \textbf{Macro} & & & & & & & \textbf{$\text{Risk}_{\text{tol}}$} & \textbf{$\text{Risk}_{\text{opt}}$} & \textbf{$\text{Risk}_{\text{av}}$} & \textbf{$\text{Risk}_{\text{tol}}$} & \textbf{$\text{Risk}_{\text{opt}}$} & \textbf{$\text{Risk}_{\text{av}}$}  & \textbf{$\text{Risk}_{\text{tol}}$} & \textbf{$\text{Risk}_{\text{opt}}$} & \textbf{$\text{Risk}_{\text{av}}$} \\
\midrule

& 3-day & 99.76 & \textbf{88.86} & \textbf{15.24} & 38.64 & 5.00 & \textbf{0.95} & \textbf{0.669} & 0.004 & \textbf{0.903} & \textbf{0.494} & 0.692 & 6.06 & 1.71 & 1.41 & 52.62 & 53.05 & 53.90 & \textbf{9.88} & 3.23 & \textbf{10.86} \\

{\textbf{GPT-5}} & 5-day & 99.29 & \textbf{87.94} & 6.43 & \textbf{32.68} & 1.19 & 0.24 & \textbf{0.727} & 0.009 & \textbf{0.906} & \textbf{0.506} & 0.864 & 7.90 & 0.51 & 1.45 & 47.44 & 46.57 & 49.28 & \textbf{6.39} & \textbf{5.59} & \textbf{8.67} \\

& 7-day & 99.52 & \textbf{87.44} & 3.11 & 34.85 & 2.38 & 0.00 & \textbf{0.748} & 0.006 & \textbf{0.911} & \textbf{0.502} & 0.948 & \textbf{7.63} & 1.67 & 1.58 & 47.71 & 48.57 & 48.93 & \textbf{7.36} & 2.43 & \textbf{11.85} \\
\midrule

& 3-day & 100.00 & 84.57 & 9.04 & \textbf{39.48} & \textbf{12.78} & 0.00 & 0.542 & 0.007 & 0.859 & 0.467 & 0.704 & 5.32 & 2.67 & 1.52 & 56.12 & 55.66 & 54.92 & 7.58 & \textbf{3.65} & 4.58 \\

\textbf{Qwen3} & 5-day & 100.00 & 83.69 & 4.28 & 32.08 & \textbf{9.16} & 0.30 & 0.580 & 0.017 & 0.826 & 0.465 & 0.876 & 11.54 & 2.42 & 2.67 & 51.89 & 52.13 & 56.39 & 1.57 & 3.77 & 6.32 \\

& 7-day & 100.00 & 82.06 & 2.74 & 31.29 & \textbf{10.45} & 0.00 & 0.632 & 0.013 & 0.880 & 0.492 & 0.949 & 8.79 & 3.26 & 3.95 & 55.19 & 53.65 & 53.22 & 3.78 & \textbf{3.98} & 2.67 \\
\midrule

\textbf{Mistral 7B} & 3-day & 100.00 & 87.81 & 10.08 & 36.95 & 2.59 & 0.00 & 0.434 & \textbf{0.072} & 0.891 & 0.291  & \textbf{0.841} & \textbf{14.99} & \textbf{7.56} & \textbf{10.03} & 53.65  & \textbf{59.17} & 56.29 & 3.28 & 0.45 & 1.14 \\

\textbf{Instruct v0.3} & 5-day & 100.00 & 87.69 & 10.31 & 30.09 & 1.22 & \textbf{0.61} & 0.436 & \textbf{0.061} & 0.811 &  0.209 & \textbf{0.929} & \textbf{28.74} & 6.58 & \textbf{15.59} & 49.03 & 52.73 & \textbf{58.11} & 1.14 & 0.97 & 0.75 \\

& 7-day & 100.00 & 78.34 & 2.67 & \textbf{36.00} & 0.00 & 0.00 & 0.365 & \textbf{0.062} & 0.805 & 0.279 & \textbf{0.975}  & \textbf{40.97} & \textbf{20.15} & \textbf{38.53} & 60.49 & \textbf{59.93} & 51.83 & 1.67 & 1.03 & 1.94 \\
\midrule

\textbf{Phi-4 mini} & 3-day & 99.22 & 87.06 & 10.27 & 35.53 & 3.16 & 0.79 & 0.193 & 0.065 & 0.659 & 0.277 & 0.809 & 8.37 & 6.70 & 9.57 & \textbf{58.31} & 58.70 & \textbf{57.08}  & 2.34 & 1.29& 2.39 \\
\textbf{Instruct} & 5-day & 92.94 & 84.97 & \textbf{11.89} & 28.66 & 0.75 & 0.38 & 0.153 & 0.059 & 0.677 & 0.224 & 0.845 & 11.67 & \textbf{7.71} & 14.56 & \textbf{57.16} & \textbf{55.58} & 56.11 & 0.74 & 0.37 & 2.41 \\
& 7-day & 92.79 & 76.12 & 3.61 & 17.32 & 0.91 & 0.00 & 0.101 & 0.055 & 0.351 & 0.054 & 0.757 & 11.05 & 5.57 & 17.86 & \textbf{63.94} & \textbf{59.93} & \textbf{53.72} & 0.74 & 1.26 & 4.75 \\
\bottomrule
\end{tabular}%
}
\caption{Combined pass-rate, qualitative and Uncertainty evaluation metrics across 3-day, 5-day, and 7-day plans. Pass-rate metrics include Delivery Rate, Commonsense Pass Rate (CPR), Hard Constraint Pass Rate (HCPR) with Micro\,/\,Macro variants, and Final Pass Rate. Qualitative evaluation metrics include: $T_{\text{meal}}$, $T_{\text{attr}}$, $S_{\text{spatial}}$, $P_{\text{persona}}$, $O_{\text{ord}}$ and BAS, CATS and TDAS metrics across different traveller type(Risk tolerance, optimizers and averse). Bolded values represent the best result among the four models.}
\label{tab:combined_metrics1}
\end{table*}

\begin{figure}[h]
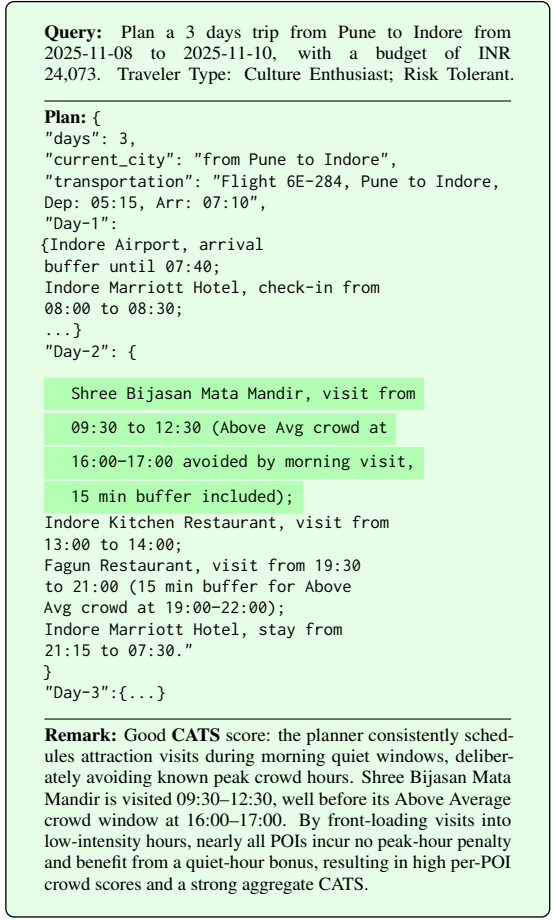

\centering
\begin{tcolorbox}[enhanced,
    colback=green!10,
    colframe=black,
    boxrule=0.5pt,
    arc=1mm,
    width=0.45\textwidth,
    fontupper=\scriptsize]
\textbf{Query:} Plan a 3 days trip from Pune to Indore from
2025-11-08 to 2025-11-10, with a budget of INR 24,073.
Traveler Type: Culture Enthusiast; Risk Tolerant.
\noindent\rule{\linewidth}{0.4pt}
\textbf{Plan:}
\texttt{\{}\\
\texttt{"days": 3,}\\
\texttt{"current\_city": "from Pune to Indore",}\\
\texttt{"transportation": "Flight 6E-284, Pune to Indore,}\\
\texttt{\hspace{1em}Dep: 05:15, Arr: 07:10",}\\
\texttt{"Day-1":}\\
\texttt{\{Indore Airport, arrival}\\
\texttt{\hspace{1em}buffer until 07:40;}\\
\texttt{\hspace{1em}Indore Marriott Hotel, check-in from}\\
\texttt{\hspace{1em}08:00 to 08:30;}\\
\texttt{...\}}\\
\texttt{"Day-2": \{}\\

\colorbox{green!30}{\texttt{\hspace{1em}Shree Bijasan Mata Mandir, visit from}}\\
\colorbox{green!30}{\texttt{\hspace{1em}09:30 to 12:30 (Above Avg crowd at}}\\
\colorbox{green!30}{\texttt{\hspace{1em}16:00--17:00 avoided by morning visit,}}\\
\colorbox{green!30}{\texttt{\hspace{1em}15 min buffer included);}}\\
\texttt{\hspace{1em}Indore Kitchen Restaurant, visit from}\\
\texttt{\hspace{1em}13:00 to 14:00;}\\
\texttt{\hspace{1em}Fagun Restaurant, visit from 19:30}\\
\texttt{\hspace{1em}to 21:00 (15 min buffer for Above}\\
\texttt{\hspace{1em}Avg crowd at 19:00--22:00);}\\
\texttt{\hspace{1em}Indore Marriott Hotel, stay from}\\
\texttt{\hspace{1em}21:15 to 07:30."}\\
\texttt{\}}\\
\texttt{"Day-3":\{...\}}\\
\noindent\rule{\linewidth}{0.4pt}
\textbf{Remark:} Good \textbf{CATS} score: the planner
consistently schedules attraction visits during morning
quiet windows, deliberately avoiding known peak crowd
hours. Shree Bijasan Mata Mandir is visited 09:30--12:30,
well before its Above Average crowd window at 16:00--17:00. By front-loading
visits into low-intensity hours, nearly all POIs incur no
peak-hour penalty and benefit from a quiet-hour bonus,
resulting in high per-POI crowd scores and a strong
aggregate CATS.
\end{tcolorbox}
\caption{Example of Crowd-Aware Scheduling Avoiding Peak-Hour Congestion}
\label{fig:case_good_cats}
\end{figure}

\subsection{Key Observations and Discussions}

\paragraph{Observation 1. Trade-offs between qualitative coherence and constraint adherence.}
Table~\ref{tab:combined_metrics1} shows distinct trade-offs between qualitative coherence and feasibility under uncertainty-aware planning. GPT-5 consistently achieves the strongest qualitative performance, leading in spatial reasoning ($S_{\text{spatial}}$), persona preservation ($P_{\text{persona}}$), and meal coherence ($T_{\text{meal}}$), while attaining the highest micro-CPR at both 3-day (88.86\%) and 7-day (87.44\%) settings. Mistral-7B-Instruct-v0.3 performs best on temporal attraction alignment ($T_{\text{attr}}$) and ordering coherence ($O_{\text{ord}}$), maintaining a perfect 100\% delivery rate across all horizons. In contrast, Qwen3 exhibits stronger hard-constraint adherence, achieving the highest macro pass rates (12.78\%, 9.16\%, and 10.45\% for 3-, 5-, and 7-day settings, respectively), whereas Phi-4-mini-Instruct shows the weakest qualitative performance with increasing degradation as planning horizon grows. These findings suggest that stronger qualitative realism does not necessarily imply stronger feasibility, revealing distinct trade-offs between coherent planning and constraint satisfaction under uncertainty.

\paragraph{Observation 2. The robustness gap exposed by uncertainty-aware metrics.}

Traditional pass-rate metrics become less informative once plans violate key constraints. Although all models maintain high delivery rates (92--100\%), final pass rates rapidly deteriorate as the planning horizon increases and approach near-zero values, making it difficult to distinguish partially robust plans from fundamentally fragile ones. In contrast, the proposed uncertainty-aware metrics (BAS, CATS, and TDAS) reveal distinct model behaviours.

For \textit{BAS}, GPT-5 and Qwen3 consistently achieve low scores (1--12\%), indicating poor temporal buffer allocation, whereas Mistral-7B-Instruct-v0.3 substantially improves with longer horizons (e.g., 14.99\%$\rightarrow$40.97\% for $\text{Risk}_{tol}$). Interestingly, $\text{Risk}_{opt}$ consistently yields the lowest BAS scores across models, suggesting that balanced scheduling with moderate buffering is more difficult than either aggressive or conservative planning.

For \textit{CATS}, all models achieve substantially higher scores (47--64\%) than BAS and TDAS, suggesting that avoiding crowded periods is comparatively easier than modeling temporal or transportation uncertainty. Phi-4-mini-Instruct achieves the strongest performance (63.94\% for $\text{Risk}_{tol}$ at 7-day), while Qwen3 consistently outperforms GPT-5; however, stronger crowd-aware behaviour does not translate into stronger overall robustness (see Example~\ref{fig:case_good_cats}).

For \textit{TDAS}, all models obtain uniformly low scores (0.37--11.85\%), substantially below human performance. GPT-5 performs best, reaching 11.85\% for $\text{Risk}_{av}$ travellers, whereas Qwen3 exhibits noticeable degradation with increasing planning horizon (7.58\%$\rightarrow$1.57\% for $\text{Risk}_{tol}$). Mistral-7B and Phi-4 remain consistently weak across traveller profiles.

Overall, uncertainty-aware metrics reveal model-specific failure patterns hidden by binary evaluation: GPT-5 and Qwen3 struggle with temporal buffering, Qwen3 degrades at longer horizons, Phi-4's strong crowd-aware behaviour masks weak temporal robustness, and all models consistently struggle with the $\text{Risk}_{opt}$ traveller profile.

\paragraph{Observation 3. Limitations in LLM-generated itineraries.}

Despite uncertainty-aware prompting, LLMs exhibit several systematic failure modes. Models frequently generate temporal inconsistencies (such as, PoI visits can extend
beyond planned departure times, meal schedules may not align with natural dining
hours, and activity timestamps can be misordered) disrupting itinerary flow (see Appendix~\ref{app:case-studies}). They also struggle to adapt activity density and temporal buffering across traveller risk profiles, as reflected by consistently low BAS scores relative to human performance. Moreover, although GPT-5 achieves stronger spatial reasoning scores ($S_{\text{spatial}}$), all models rely on optimistic transit assumptions in multi-city settings, leading to fragile itineraries under realistic delays.

These limitations become more prominent with increasing planning horizon. Performance degrades from 3-day to 7-day itineraries, while human-annotated plans also decline from 91\% to 63\%, indicating that long-horizon planning is inherently challenging even for humans. Nevertheless, model performance remains substantially below human levels across all settings, suggesting that current LLMs still struggle to reason effectively over stochastic travel dynamics, user risk preferences, and long-range temporal dependencies.

\subsection{Agentic baseline.}
To evaluate the effectiveness of explicit uncertainty modeling beyond
agentic planning, we additionally compare against ATLAS~\cite{choi2026atlas},
an agentic framework for real-world travel planning. Table~\ref{tab:combined_metrics} provides a
complementary comparison between agentic task decomposition and our
uncertainty-aware planning approach.

\begin{table*}[t]
\centering
\footnotesize
\setlength{\tabcolsep}{3pt}
\renewcommand{\arraystretch}{1.15}
\caption{Combined pass-rate and uncertainty metrics for Phi-4-mini across
different traveler types (risk-tolerant, risk-optimistic, and risk-averse)
for 3-day, 5-day, and 7-day plans using agentic baseline ATLAS\cite{choi2026atlas}.}
\label{tab:combined_metrics}

\resizebox{\textwidth}{!}{%
\begin{tabular}{@{}l
  S[table-format=3.2]
  S[table-format=3.2]
  S[table-format=2.2] S[table-format=2.2]
  S[table-format=2.2] S[table-format=2.2]
  S[table-format=2.2]
  S[table-format=2.2] S[table-format=2.2] S[table-format=2.2]
  S[table-format=2.2] S[table-format=2.2] S[table-format=2.2]
  S[table-format=2.2] S[table-format=2.2] S[table-format=2.2]@{}}

\toprule
& &
\multicolumn{2}{c}{\textbf{CPR}} &
\multicolumn{2}{c}{\textbf{HCPR}} &
&
\multicolumn{3}{c}{\textbf{BAS}} &
\multicolumn{3}{c}{\textbf{CATS}} &
\multicolumn{3}{c}{\textbf{TDAS}} \\

\cmidrule(lr){3-4}
\cmidrule(lr){5-6}
\cmidrule(lr){8-10}
\cmidrule(lr){11-13}
\cmidrule(lr){14-16}

\textbf{Type} &
{\textbf{Del}} &
{\textbf{Micro}} &
{\textbf{Macro}} &
{\textbf{Micro}} &
{\textbf{Macro}} &
{\shortstack{\textbf{Final}\\\textbf{Pass Rate}}} &
{$Risk_{tol}$} &
{$Risk_{opt}$} &
{$Risk_{av}$} &
{$Risk_{tol}$} &
{$Risk_{opt}$} &
{$Risk_{av}$} &
{$Risk_{tol}$} &
{$Risk_{opt}$} &
{$Risk_{ave}$} \\

\midrule

3d &
100.00 & 87.31 & 8.93 & 40.31 & 40.32 & 5.36 &
15.23 & 6.57 & 5.00 &
51.72 & 50.65 & 51.51 &
1.34 & 2.64 & 3.12 \\

5d &
100.00 & 82.83 & 6.74 & 15.55 & 15.55 & 1.82 &
25.34 & 10.54 & 10.04 &
50.62 & 54.99 & 55.52 &
0.85 & 0.42 & 2.84 \\

7d &
99.74 & 78.88 & 7.02 & 0.51 & 0.53 & 0.00 &
27.57 & 15.40 & 13.03 &
51.98 & 51.08 & 52.33 &
0.00 & 0.00 & 0.87 \\

\bottomrule
\end{tabular}%
}
\end{table*}
\paragraph{Analysis.}
ATLAS does not consistently outperform our uncertainty-aware prompting
strategy despite its agentic planning framework. While it achieves
competitive performance on selected crowd-awareness measures such as
CATS, its performance is lower on delivery rate, constraint preservation,
and several uncertainty-aware metrics, particularly BAS and TDAS for
longer itineraries. These results indicate that agentic decomposition
alone does not fully address the challenges posed by UTP-Bench and
highlight the importance of explicitly modeling travel uncertainty.

\subsection{Performance Without Uncertainty}

We also evaluated GPT-5 on the UTP-Bench dataset without explicitly
modeling uncertainty factors, such as flight delays, traffic, and weather
disruptions. This setting provides a deterministic baseline for assessing
the model's ability to satisfy travel planning constraints before
introducing real-world uncertainty. Results are reported in
Table~\ref{tab:qualitative_metrics} and~\ref{tab:pass_rates}. 

The qualitative metrics remain relatively stable across itinerary lengths.
The meal-related score ($T_{\text{meal}}$) increases from 0.66 for 3-day
plans to 0.74 for 7-day plans, while spatial consistency ($S_{\text{spat}}$)
remains within 0.89--0.91. Persona alignment ($P_{\text{pers}}$) remains
stable at approximately 0.47--0.48. The ordering score ($O_{\text{ord}}$)
increases from 0.73 to 0.95 with increasing plan length, indicating
stronger ordering consistency. The $T_{\text{attr}}$ values remain low
across all settings (0.0057--0.0087), consistent with the metric
definition, where lower values indicate better performance.
Despite these strong qualitative scores, constraint-based metrics show
clear degradation as itinerary length increases. Micro-CPR decreases from
84\% for 3-day plans to 76\% and 71\% for 5- and 7-day plans,
respectively, while macro-CPR decreases from 29\% to 14\% and 4.6\%.
Similarly, macro-HCPR decreases from 10.66\% to 2.13\% and 0\%, while
the Final Pass Rate falls from 2.5\% to 0.677\% and 0\% across 3-, 5-,
and 7-day plans.

The divergence between qualitative scores and constraint satisfaction
highlights an important limitation of LLM-based travel planning: an
itinerary can appear coherent and preference-aligned while violating
underlying constraints. As itinerary length increases, interacting
temporal, spatial, and planning constraints become harder to satisfy
simultaneously. The zero Final Pass Rate for 7-day plans demonstrates
this compounding effect even without explicit uncertainty, motivating
uncertainty-aware planning in UTP-Bench.

\begin{table}[]
\centering
\caption{Qualitative metrics and Delivery rates for GPT-5 (3,5,7-day plans).}
\label{tab:qualitative_metrics}
\resizebox{\columnwidth}{!}{
\begin{tabular}{l c c c c c c c}
\toprule
\textbf{Type} & $T_{\text{meal}}$ & $T_{\text{attr}}$ & $S_{\text{spat}}$ & $P_{\text{pers}}$ & $O_{\text{ord}}$ & \textbf{Deliv.} \\
& & & & & & \textbf{Rate} \\
\midrule
3-day & 0.66 & 0.0087 & 0.89 & 0.48 & 0.73 & 94.1 \\
5-day & 0.72 & 0.0057 & 0.90 & 0.47 & 0.91 & 98.7 \\
7-day & 0.74 & 0.0063 & 0.91 & 0.48 & 0.95 & 96.8 \\
\bottomrule
\end{tabular}
}
\end{table}

\begin{table}[]
\centering
\caption{Pass-rates (CPR/HCPR) and Final Pass for GPT-5 (3,5,7-day plans).}
\label{tab:pass_rates}
\resizebox{\columnwidth}{!}{
\begin{tabular}{l c c c c c}
\toprule
\textbf{Type} & \multicolumn{2}{c}{\textbf{CPR (\%)}} & \multicolumn{2}{c}{\textbf{HCPR (\%)}} & \textbf{Final} \\
\cmidrule(lr){2-3} \cmidrule(lr){4-5}
& \textbf{Micr.} & \textbf{Macr.} & \textbf{Micr.} & \textbf{Macr.} & \textbf{Pass} \\
\midrule
3-day & 84 & 29 & 16.3 & 10.66 & 2.5 \\
5-day & 76 & 14 & 6.7 & 2.13 & 0.677 \\
7-day & 71 & 4.6 & 7.9 & 0.0 & 0.0 \\
\bottomrule
\end{tabular}
}
\end{table}

\section{Conclusion}

We introduced \system{}, a high-fidelity benchmark for uncertainty-aware travel itinerary planning that bridges the gap between deterministic planning assumptions and the stochastic nature of real-world travel. By incorporating fine-grained attraction categories, traveller risk profiles, and empirical transportation-delay and crowd-dynamics signals, \system{} provides a realistic testbed for evaluating spatio-temporal reasoning in LLM-based itinerary generation. We further proposed three uncertainty-aware evaluation metrics namely \textit{BAS, CATS and TDAS} to assess itinerary robustness beyond conventional constraint-satisfaction metrics. Experimental results demonstrate that although current LLMs can generate qualitatively coherent itineraries, they struggle to reason effectively over temporal uncertainty, transportation variability, traveller risk preferences, and long-horizon dependencies. Overall, \system{} advances travel-planning evaluation from deterministic feasibility to robustness under realistic uncertainty, enabling more reliable and adaptive human-centric planning systems.

\section*{Limitations}
While \system{} significantly enhances the realism and coherence of travel
planning benchmarks, certain limitations remain. 
First, \system{} is designed as a static benchmark and relies on historical statistics rather than real-time signals. Incorporating live transportation and traffic APIs represents a promising direction for future extensions.
Second, our objective is not to propose a novel planning architecture, but rather to establish a benchmark and uncertainty-aware evaluation framework for robust travel planning. \system{} provides a strong foundation for assessing LLM-driven travel planning, future research may explore diverse planning paradigms, including agentic and tool-augmented systems, and evaluate their ability to reason under realistic travel uncertainty.
Another key constraint is the geographic scope, which is currently concentrated on Indian. If the necessary data becomes available, the construction pipeline can be extended to any regions, enabling broader geographic generalization.  Our benchmark is currently designed primarily in English, which does not fully capture the multilingual nature of the Indian travel ecosystem. Expanding to regional languages would require accounting for language-specific differences in place names, cultural travel preferences, and transit terminology, which remain open challenges for future research.

\section*{Ethical Considerations}
Our study utilizes publicly available data from the Internet, which we have carefully scraped to construct our databases while ensuring compliance with relevant terms of use and ethical considerations. To safeguard privacy, we have fully anonymized sensitive personal details. However, with annotators consent, aggregate demographic statistics are provided in the Appendix ~\ref{app:annotator-details}. AI-assisted tools were used during the preparation of this work for language refinement and polishing. All research decisions, methodology, data curation, experiments, analyses, and conclusions were performed and verified by the authors. 

\section*{Acknowledgments}
This research was partially supported by the Technology Innovation Hub (TIH), IIT Tirupati (IITTNiF/TPD/2024-25/P16). We sincerely thank the annotators for their efforts in annotating, validating, and quality-checking the dataset. We also thank the anonymous reviewers for their valuable comments and constructive feedback, which helped improve this work.

\bibliography{references}

\appendix

\vspace{0.5cm}
\newpage
\noindent {\Large \textbf{Overview of Appendix Sections}}

\begin{itemize}
  \item Appendix~\ref{app:data-sourcing}: Data Sourcing Details
  \item Appendix~\ref{app:more-evaluation}: More about Evaluation
  % \item Appendix~\ref{app:delay-distr}: Uncertainty Distribution for Attraction and Restaurants
  \item Appendix~\ref{app:prompts}: Prompt and Annotation Details
  \item Appendix~\ref{app:case-studies}: Case Studies
  \item Appendix~\ref{app:annotator-details}: Annotator Guidelines and Demographics
\end{itemize}

\section{Data Sourcing Details}
\label{app:data-sourcing}

UTP-Bench is constructed entirely from real-world, India-specific data sources to ensure spatio-temporal consistency, operational accuracy, and uncertainty-awareness across all database components. Below, we detail the sourcing methodology and heuristics for each component.

\subsection{Restaurants}
Restaurant details were extracted using the TripAdvisor API\footnote{\url{https://www.tripadvisor.com/}}, providing attributes including name, category, cuisine type, ratings, reviews, and price tier.The cost is denoted using the symbol (\$\$\$) rather than exact values. To estimate absolute prices, we leveraged city-specific restaurant price indices from Numbeo\footnote{\url{https://www.numbeo.com/cost-of-living/}}, scaling them according to the price tier rating of each restaurant entry. Additionally, crowd density and peak visiting hours for restaurants were collected via the BestTime API\footnote{\url{https://besttime.app}}, providing hour-by-hour crowd intensity data across 20 major Indian cities. This data forms the empirical foundation for the crowd-induced delay modeling in UTP-Bench's CATS and BAS uncertainty metrics.

\subsection{Attractions}
Attraction details, including subcategories and ratings, were sourced from the TripAdvisor API.\footnote{\url{https://www.tripadvisor.com/}} Since a majority of attractions lacked predefined visit durations, we adopted the category-wise average durations from TripCraft, assigning each attraction's duration as the mean of the categories it belonged to, ensuring a realistic time allocation (Table~\ref{tab:full_attraction_durations}). Crowd density and peak-hour patterns for attractions were collected via the BestTime API, covering 20 major Indian cities and enabling crowd-sensitive scheduling through the CATS metric. All 504 cities in the UTP-Bench city set were validated against attraction availability, with cities lacking sufficient PoI coverage filtered out during the data cleaning phase.

\subsection{Accommodations}
Accommodation listings were collected via the TripAdvisor API,\footnote{\url{https://www.tripadvisor.com/}} providing attributes including name, property type, star rating, amenities, and location coordinates. Absolute nightly prices were calibrated using city-level accommodation cost baselines from Numbeo,\footnote{\url{https://www.numbeo.com/cost-of-living/}} ensuring budget constraint evaluations are grounded in realistic India-specific pricing. Accommodation listings are used as the designated start and end point for each day's PoI sequence in UTP-Bench's itinerary structure.

\subsection{Trains}
Train schedules were sourced from Tripozo,\footnote{\url{https://www.tripozo.com/}} yielding 68{,}068 train records covering intercity and interstate routes across India. Each record includes train number, source and destination stations, departure and arrival times, journey duration, and availability by class. Historical delay data for trains was collected across multiple time horizons (1 week, 1 month, 3 months, 6 months, and 1 year), providing the empirical delay distributions used in UTP-Bench's TDAS metric for train delay absorption modeling.

\subsection{Buses}
Intercity bus schedules were sourced from Tripozo,\footnote{\url{https://www.tripozo.com/}}, yielding 34{,}925 bus records covering state and private operator routes across India. Each record includes operator name, source and destination cities, departure and arrival times, bus type, and fare. Road delay information for bus routes was obtained from the TomTom API,\footnote{\url{https://developer.tomtom.com/}} covering 4{,}052 city-pair routes with empirical traffic delay distributions, providing the delay parameters used in TDAS bus delay modeling.

\subsection{Flights}
Domestic flight schedules were sourced from Tripozo,\footnote{\url{https://www.tripozo.com/}} yielding 29{,}346 flight records across 67 Indian cities served by domestic carriers. Each record includes airline, flight number, source and destination airports, departure and arrival times, and fare class.Delay information is collected via FlightStats \footnote{https://developer.flightstats.com/}. For each flight, on-time performance statistics including on-time probability $p_{\text{on-time}}$, mean delay $\mu_{\text{late}}$, and standard deviation $\sigma_{\text{late}}$ were derived from historical records, forming the flight delay distribution parameters used in UTP-Bench's TDAS metric.

\subsection{Events}
Event data was scraped from District by Zomato,\footnote{\url{https://district.zomato.com/}} covering a diverse range of concerts, sports, arts \& theatre, music, and film events across India. The scraped dataset yields 783 events.

\subsection{Road Delay and Traffic Data}
City-to-city road delay information was collected via the TomTom Traffic API,\footnote{\url{https://developer.tomtom.com/}} covering 4{,}052 city-pair routes across India. Each route entry includes Estimated delay, Historical delay
and Free Flow Traffic.

\subsection{Crowd Density and Best-Time Data}
Peak-hour crowd density patterns for attractions and restaurants were collected via the BestTime API,\footnote{\url{https://besttime.app}} which provides hour-by-hour crowd intensity classifications across a week for each venue. Data was collected for venues across 20 major Indian cities, forming the empirical backbone of UTP-Bench's crowd-induced delay modeling. Crowd intensity is classified into five tiers (Low, Below Average, Average, Above Average, and High) with corresponding congestion factors $C_f$ used in BAS and risk values $R(t)$ used in CATS.

\subsection{Cost of Living}
City-level cost-of-living data was sourced from Numbeo,\footnote{\url{https://www.numbeo.com/cost-of-living/}} covering 20 major Indian cities. Attributes include average meal costs across price tiers, local transport fares, and accommodation price baselines. These data points are used for two purposes: (1) calibrating budget constraints in travel queries, and (2) converting relative price tier ratings from TripAdvisor into absolute cost estimates for restaurants and accommodations. 

\subsection{Distance Matrix}
All pairwise city-to-city distances and travel times were computed using the TomTom Routing API\footnote{\url{https://developer.tomtom.com/}} covering 253{,}513 city-pair entries across UTP-Bench's 504-city set. For each Point of Interest, including accommodations, restaurants, and attractions, the nearest public transit stop was determined using geo distance computed via OpenStreetMap API\footnote{\url{https://www.openstreetmap.org/}}, yielding 10{,}025 nearest transit stop entries. This enables realistic spatial reasoning and transit-aware itinerary generation across India's diverse urban and semi-urban geographies.

\begin{table}[t]
\centering
\scriptsize
\caption{Attraction visiting duration (hrs) for each category. Note that an attraction can belong to one or more categories.}
\label{tab:full_attraction_durations}
\begin{tabular}{l r}
\toprule
Category & Duration (hrs) \\
\midrule
Boat Tours \& Water Sports & 3.5 \\
Casinos \& Gambling & 2.5 \\
Classes \& Workshops & 1.5 \\
Concerts \& Shows & 2.5 \\
Food \& Drink & 2.5 \\
Fun \& Games & 1.5 \\
Museums & 3.0 \\
Nature \& Parks & 4.5 \\
Nightlife & 2.5 \\
Outdoor Activities & 4.0 \\
Shopping & 1.5 \\
Sights \& Landmarks & 3.0 \\
Spas \& Wellness & 2.0 \\
Water \& Amusement Parks & 5.0 \\
Zoos \& Aquariums & 2.5 \\
\bottomrule
\end{tabular}
\end{table}

\begin{table*}[t]
\centering
\scriptsize
\caption{Comprehensive constraint and persona description. }
\label{tab:full_constraints}
\begin{tabular}{p{0.28\linewidth} p{0.66\linewidth}}
\toprule
\textbf{Constraint} & \textbf{Description} \\
\midrule
\multicolumn{2}{l}{\textbf{Environment Constraint}} \\
Unavailable Transportation & There is no available flight or driving information between the two cities. \\
Unavailable Attractions & There is no available attraction information in the queried city. \\
\midrule
\multicolumn{2}{l}{\textbf{Commonsense Constraint}} \\
Within Sandbox & All information in the plan must be within the closed sandbox; otherwise, it will be considered a hallucination. \\
Complete Information & No key information should be left out of the plan, such as the lack of accommodation during travel. \\
Sufficient Meal Gaps & Meal timings must have a minimum gap of four hours between breakfast, lunch, and dinner to maintain a natural schedule. \\
Valid PoI list  & The point-of-interest (PoI) list must follow strict validity rules: each day's itinerary must begin and end at the designated accommodation, except on the final day when the traveler departs. The list should be limited to accommodations, attractions, and restaurants, ensuring adequate time gaps between flight arrivals and accommodation check-ins, as well as between accommodation check-outs and departures. \\
Diverse Events  & Event choices should not be repeated throughout the trip. \\
Within Current City & All scheduled activities for the day must be located within that day's city(ies). \\
Reasonable City Route & Changes in cities during the trip must be reasonable. \\
Diverse Restaurants & Restaurant choices should not be repeated throughout the trip. \\
Diverse Attractions & Attraction choices should not be repeated throughout the trip. \\
Non-conf. Transportation & Transportation choices within the trip must be reasonable. For example, having both ``cab/taxi'' and ``flight'' would be considered a conflict. \\
\midrule
\multicolumn{2}{l}{\textbf{Hard Constraint}} \\
Budget & The total budget of the trip. \\
House Rule & House rules include ``Safety Conduct'', ``No smoking'', ``No children'', ``No pets'', and ``Checkin/Checkout''. \\
House Type & House types include ``Apartment Extended'', ``Luxury Boutique'', ``Resort/Beach'', and ``Family Friendly''. \\
Cuisine Group &
Cuisines include "African", "American", "Asian", "Bar/Pub/Cafe", "Eastern European/Russian", "European", "Indian Subcontinent", "International Fusion", "Latin/South American", "Mediterranean/Middle Eastern", "Pizza/Italian Specialties", and "Seafood/Steak/Grill". \\
Transportation & Transportation options include ``No flight'' and ``No Cab''. \\
Attraction Types & Each attraction belongs to one or more of 15 predefined categories, ensuring a well-distributed selection of activities. \\
\midrule
\multicolumn{2}{l}{\textbf{Persona Components}} \\
Traveler Type & Defines how a traveler approaches their journey  whether they seek relaxation in cozy spots or adrenaline-pumping adventures. \\
Purpose of Travel & Captures the main motivation behind the trip, whether it is to unwind, seek thrills, explore cultures, or connect with nature. \\
Spending Preference & Reflects the traveler's budget and style, from luxurious indulgence to cost-conscious experiences. \\
Location Preference & Highlights preferred environments, such as beaches, mountains, cities, or wildlife-rich forests. \\
\midrule
\multicolumn{2}{l}{\textbf{Uncertainty Constraints $\dagger$}} \\
Transit Delay Buffer & Buffers must account for empirical delay distributions across flights, trains, buses, and cabs. \\
Crowd-Induced Delay & Visit and wait times must reflect peak-hour crowd density patterns. \\
Activity Buffer Gap & Gaps between activities must absorb real-world transit and crowd disruptions. \\
\midrule
\multicolumn{2}{l}{\textbf{Persona Components}} \\
Traveler Type & Laid-back or adventure-seeking travel style. \\
Purpose of Travel & Relaxation, adventure, cultural exploration, or nature. \\
Spending Preference & Luxury or budget-conscious, shaping all cost decisions. \\
Location Preference & Beaches, mountains, heritage cities, wildlife, or religious destinations. \\
Risk Profile $\dagger$ & Risk-Tolerant, Risk Optimized, or Risk-Averse calibrates uncertainty tolerance. \\
\bottomrule
\end{tabular}

\end{table*}

\section{Evaluation Metrics (Details)}
\label{app:more-evaluation}

 We introduce four complementary metrics to evaluate the temporal realism, crowd-awareness, transport robustness, and global schedule resilience of travel itineraries.  Each metric is computed independently for human-annotated and LLM-generated plans.

\subsection{Buffer Adequacy Score (BAS)}

\paragraph{Overview.}
\noindent\textbf{Buffer Adequacy Score (BAS)} measures the temporal realism and feasibility of a travel itinerary by evaluating whether the buffer time
allocated at each Point of Interest (PoI) is consistent with the base visit
duration, scaled appropriately to crowd congestion, and within the range
expected for the traveller's risk profile.

\paragraph{Inputs.}
For each PoI the following inputs are required:

\begin{itemize}
  \item Place name and type (\textit{Attraction} or \textit{Restaurant})
  \item Arrival time $t_a$ and departure time $t_d$ (minutes from midnight)
  \item Base visit duration $v$ (minutes), defined as follows:
\end{itemize}

\noindent
For \textbf{Attractions}, $v$ is taken from the \texttt{visit\_duration}
column of the attractions dataset.
For \textbf{Restaurants}, $v$ is derived from the meal type determined by
start hour, as shown in Table~\ref{tab:meal_duration}.

\begin{center}
\scriptsize
\begin{tabular}{lcc}
\toprule
\textbf{Meal Type} & \textbf{Start Hour} & \textbf{Duration $v$} \\
\midrule
Breakfast & 06:00 -- 10:59 & 50 min \\
Lunch     & 11:00 -- 15:59 & 60 min \\
Dinner    & 16:00 -- 23:59 & 75 min \\
\bottomrule
\end{tabular}
\captionof{table}{Base visit duration $v$ for restaurants by meal type.}
\label{tab:meal_duration}
\end{center}

\noindent
The congestion tier factor $C_f$ is parsed from the crowd description field
(\texttt{poi\_str}) and mapped as shown in Table~\ref{tab:congestion}.

\begin{table}[H]
\centering
\scriptsize

\centering
\begin{tabular}{lcc}
\toprule
\textbf{Crowd Level} & \textbf{$C_f$} \\
\midrule
Low           & 1.00 \\
Below Average & 1.25 \\
Average       & 1.50 \\
Above Average & 1.75 \\
High          & 2.00 \\
\bottomrule
\end{tabular}
\captionof{table}{Congestion tier factor $C_f$ derived from BestTime API crowd analysis.}
\label{tab:congestion}
% \end{minipage}
\end{table}

\noindent
The traveller risk-profile multipliers $[\mathit{low\_mult},\,\mathit{high\_mult}]$
are defined in Table~\ref{tab:risk_profile_bas}.

\begin{table}[H]
\centering
\scriptsize
\begin{tabular}{lcc}
\toprule
\textbf{Traveller Type} & \textbf{$\mathit{low\_mult}$} & \textbf{$\mathit{high\_mult}$} \\
\midrule
Risk Tolerant  & 0.50 & 1.00 \\
Risk Optimized & 1.00 & 1.25 \\
Risk Averse    & 1.25 & 2.00 \\
\bottomrule
\end{tabular}
\caption{Risk-profile buffer multipliers for BAS.}
\label{tab:risk_profile_bas}
\end{table}

\paragraph{Step 1: Actual Buffer.}
The actual buffer is the free time allocated by the planner beyond the base
visit duration:

\begin{equation}
  B_{\text{actual}} = (t_d - t_a) - v
\end{equation}

\noindent
$B_{\text{actual}}$ captures the \emph{extra time padded by the planner}
on top of the standard visit duration; it is not travel time to the next place.

\paragraph{Step 2: Expected Buffer Range.}
The expected buffer range is derived from the base visit duration, the
congestion factor, and the traveller risk profile with no hardcoded
category thresholds:
\begin{equation}
  a_c = \mathit{low\_mult} \times v \times C_f
\end{equation}
\begin{equation}
  b_c = \mathit{high\_mult} \times v \times C_f
\end{equation}
\noindent
This ensures that a high-crowd PoI ($C_f = 2.0$) naturally demands a larger
buffer, and a risk-averse traveller requires a proportionally wider range.
Both effects are fully data-driven.

\paragraph{Step 3: Quantitative Deviation Score.}
The distance of the actual buffer from the expected range, normalised by
the range width:
\begin{equation}
\scriptsize
  D =
  \begin{cases}
    0,
      & B_{\text{actual}} \in [a_c,\, b_c] \\[6pt]
    \min\!\left(|B_{\text{actual}} - a_c|,\;
                |B_{\text{actual}} - b_c|\right),
      & \text{otherwise}
  \end{cases}
\end{equation}
\begin{equation}
  Q = \frac{D}{b_c - a_c}
\end{equation}

\paragraph{Step 4: Final BAS.}
For an itinerary with $N$ PoI visits:
\begin{equation}
  \boxed{
    \mathrm{BAS} = 1 - \frac{1}{N}\sum_{i=1}^{N} min(1,Q_i)
  }
\end{equation}

\paragraph{Interpretation.}
\begin{itemize}
  \item $\mathrm{BAS} \approx 1$: Buffer is consistent with visit duration,
        crowd level, and traveller risk profile.
  \item $\mathrm{BAS} \approx 0$: Buffer is too short or too long relative
        to expected planning behaviour.
\end{itemize}

\subsection{Crowd-Aware Timing Score (CATS)}
\label{app:cats}
\paragraph{Overview.}
\textbf{Crowd-Aware Timing Score (CATS)} evaluates how well an itinerary
aligns visit timings with crowd patterns at each PoI, using place-specific
crowd-window data obtained from the BestTime API.
CATS does \emph{not} use PoI category or planner-assigned visit duration;
scores are driven entirely by crowd data, making the metric fully independent
of subjective category labels.

\paragraph{Inputs.}
For each PoI:
\begin{itemize}
  \item Place name
  \item Visit start time $t_{\text{start}}$ and end time $t_{\text{end}}$
  \item \texttt{avg\_besttime\_analysis} containing:
    \begin{itemize}
      \item \texttt{cw}: crowd windows: $[\mathit{start\_hour},
            \mathit{end\_hour}, \mathit{level}]$
      \item \texttt{q}: quiet hours list
      \item \texttt{p}: peak hours list
    \end{itemize}
\end{itemize}

\paragraph{Crowd Intensity Mapping.}
Each crowd level is mapped to a numeric intensity using a uniform ordinal
scale with equal intervals of 0.25 (Table~\ref{tab:crowd_intensity}).

\begin{table}[H]
\centering
\scriptsize
\begin{tabular}{llc}
\toprule
\textbf{Label} & \textbf{Meaning}    & \textbf{Intensity} \\
\midrule
L  & Low           & 0.00 \\
BA & Below Average & 0.25 \\
A  & Average       & 0.50 \\
AA & Above Average & 0.75 \\
H  & High          & 1.00 \\
\bottomrule
\end{tabular}
\caption{Crowd level to numeric intensity mapping for CATS.}
\label{tab:crowd_intensity}
\end{table}

\paragraph{Step 1: Weighted Crowd Intensity.}
A visit may span multiple crowd windows.
Intensity is computed as a weighted average proportional to actual
overlap minutes:
\begin{equation}
  \mathit{intensity} =
    \frac{\displaystyle\sum_{w} \mathit{overlap}_w \times \mathit{intensity}_w}
         {\mathit{visit\_duration}}
\end{equation}
\noindent
where $\mathit{overlap}_w$ is the number of minutes the visit overlaps
crowd window $w$.
This ensures that a visit spanning both a Below Average and an Above Average
window is scored proportionally rather than on the start hour alone.

\paragraph{Step 2: Base Score.}
\begin{equation}
  \mathit{base\_score} = 1 - \mathit{intensity}
\end{equation}

\paragraph{Step 3: Quiet-Hour Bonus.}
Overlap with quiet hours $q$ is measured proportionally:

\begin{equation}
  \mathit{quiet\_ratio} =
    \frac{\displaystyle\sum_{h \in q}
          \mathit{overlap}_h}
         {\mathit{visit\_duration}}
\end{equation}

\noindent
The bonus scales with how uncrowded the place already is:

\begin{equation}
  \mathit{quiet\_bonus} = \mathit{quiet\_ratio} \times (1 - \mathit{intensity})
\end{equation}

\noindent
A higher quiet overlap combined with a lower intensity yields a larger reward.
If \texttt{q} is absent, $\mathit{quiet\_bonus} = 0$.

\paragraph{Step 4: Peak-Hour Penalty.}
Overlap with peak hours $p$ is measured proportionally:

\begin{equation}
  \mathit{peak\_ratio} =
    \frac{\displaystyle\sum_{h \in p}
          \mathit{overlap}_h}
         {\mathit{visit\_duration}}
\end{equation}

\noindent
The penalty scales with how crowded the place is:

\begin{equation}
  \mathit{peak\_penalty} = \mathit{peak\_ratio} \times \mathit{intensity}
\end{equation}

\noindent
A higher peak overlap combined with a higher intensity yields a larger penalty.
If \texttt{p} is absent, $\mathit{peak\_penalty} = 0$.

\paragraph{Step 5: Per-PoI Score.}

\begin{eqnarray}
  \nonumber\mathit{score}_i &=& \mathrm{clip}(\mathit{base} + \mathit{quiet\_bonus}\\ 
  &-& \mathit{peak\_penalty},\ 0,\ 1)
\end{eqnarray}
\noindent
Case handling based on data availability is summarised in
Table~\ref{tab:cats_cases}.

\begin{table}[H]
\centering
\scriptsize
\begin{tabular}{lccp{4cm}}
\toprule
\textbf{Case}    & \textbf{\texttt{q}} & \textbf{\texttt{p}} & \textbf{Effect} \\
\midrule
\texttt{full}       & \checkmark & \checkmark &  bonus and penalty applied \\
\texttt{peak\_only} & $\times$   & \checkmark & Only penalty applied \\
\texttt{quiet\_only}& \checkmark & $\times$   & Only bonus applied \\
\texttt{cw\_only}   & $\times$   & $\times$   & Intensity-based score only \\
\texttt{no\_data}   & ---        & ---        & PoI skipped from scoring \\
\bottomrule
\end{tabular}
\caption{CATS case handling based on available crowd data fields.}
\label{tab:cats_cases}
\end{table}

\paragraph{Step 6: Aggregate CATS.}
For an itinerary with $N$ valid PoI visits:

\begin{equation}
  \boxed{
    \mathrm{CATS} = \frac{1}{N}\sum_{i=1}^{N} \mathit{score}_i
  }
\end{equation}

\paragraph{Interpretation.}
\begin{itemize}
  \item $\mathrm{CATS} \approx 1$: Visits are well-timed around crowd patterns.
  \item $\mathrm{CATS} \approx 0$: Visits clash heavily with peak crowd hours.
\end{itemize}

\subsection{Transport Delay Absorption Score (TDAS)}

\paragraph{Overview.}
\textbf{Transport Delay Absorption Score (TDAS)} measures whether the buffer
embedded in each planned transport segment is sufficient to absorb historical
delays across flights, trains, buses, and cabs without disrupting the overall
itinerary.

\paragraph{Inputs.}
For each transport segment $j$:
\begin{itemize}
  \item Transportation mode
    $m_j \in \{\text{Flight},\,\text{Train},\,\text{Cab},\,\text{Bus}\}$
  \item Planned duration from the itinerary (departure $\to$ arrival as
        written by the planner)
  \item Historical duration from the transport dataset (actual
        departure $\to$ actual arrival)
  \item \texttt{OnTimePerformance (JSON)} and \texttt{DelayStats (JSON)}
        for flights
  \item Multi-horizon average delay columns for trains
  \item Free-flow travel time and historical road delay for cabs/buses
  \item Traveller risk profile
\end{itemize}

\paragraph{Step 1: Buffer Extraction.}
The buffer is already embedded in the planned itinerary by the planner:

\begin{equation}
  B_j = \mathit{planned\_duration}_j - \mathit{hist\_duration}_j
\end{equation}

\noindent
where $\mathit{planned\_duration}$ is the arrival minus departure time as
given in the plan (including planner-added buffer) and
$\mathit{hist\_duration}$ is the actual historical duration from the
transport dataset.

\paragraph{Step 2: Expected Delay $E[D]$ by Transport Mode.}

\noindent\textbf{Flight --- Robust Delay Formula.}
The expected delay for flights accounts for high-variance delays and
worst-case disruptions:

\begin{equation}
  E[D]_{\text{flight}} =
    \frac{n_{\text{late}}}{N}\,(\mu + \sigma)
    \;+\;
    \frac{n_{\text{dis}}}{N}\,D_{\max}
\end{equation}

\noindent
where the symbols are defined in Table~\ref{tab:flight_symbols}.

\begin{table}[H]
\centering
\scriptsize
\begin{tabular}{cl}
\toprule
\textbf{Symbol}         & \textbf{Meaning} \\
\midrule
$N$                     & \texttt{total\_flights} \\
$n_{\text{late}}$       & \texttt{late} $+$ \texttt{very\_late} $+$ \texttt{excessive} \\
$n_{\text{dis}}$        & \texttt{cancelled} $+$ \texttt{diverted} \\
$\mu$                   & \texttt{avg\_delay\_min} \\
$\sigma$                & \texttt{std\_delay\_min} \\
$D_{\max}$              & \texttt{max\_delay\_min} \\
\bottomrule
\end{tabular}
\caption{Symbols used in the flight robust delay formula.}
\label{tab:flight_symbols}
\end{table}

\noindent
The term $\mu + \sigma$ captures high-variance delays at approximately the
84th percentile under a normal distribution.
The term $\frac{n_{\text{dis}}}{N} \times D_{\max}$ penalises worst-case
disruptions proportionally.

\medskip
\noindent\textbf{Train --- Weighted Historical Delay Mean.}
A weighted average across multiple time horizons captures structural delay
patterns, with longer horizons receiving greater weight:

\begin{equation}
  E[D]_{\text{train}} =
    \frac{d_{1w} + 2d_{1m} + 3d_{3m} + 4d_{6m} + 6d_{1y}}{16}
\end{equation}

\noindent
where $d_h$ is the average delay over horizon $h$
(1 week, 1 month, 3 months, 6 months, 1 year).
Recent short horizons are weighted less; longer horizons reflect structural
delay patterns in the rail network.

\medskip
\noindent\textbf{Cab / Bus --- Empirical Road Delay.}

\begin{equation}
  E[D]_{\text{road}} = \text{Est. historical delay from road dataset}
\end{equation}

\begin{equation}
  \mathit{hist\_duration} = \text{Free-flow travel time from road dataset}
\end{equation}

\noindent
Empirical road-delay distributions are sourced from the TomTom API.

\paragraph{Step 3: Risk-Profile Aware Expected Buffer Range.}
Each traveller risk profile maps to a safe buffer multiplier range
(Table~\ref{tab:risk_profile_tdas}):

\begin{table}[H]
\centering
\scriptsize
\begin{tabular}{lcc}
\toprule
\textbf{Traveller Type} & \textbf{$\mathit{low\_mult}$} & \textbf{$\mathit{high\_mult}$} \\
\midrule
Risk Tolerant  & 0.50 & 1.00 \\
Risk Optimized & 1.00 & 1.25 \\
Risk Averse    & 1.25 & 2.00 \\
\bottomrule
\end{tabular}
\caption{Risk-profile buffer multipliers for TDAS.}
\label{tab:risk_profile_tdas}
\end{table}

\noindent
The expected safe buffer range for segment $j$ is:

\begin{equation}
  a_c = \mathit{low\_mult} \times E[D]
\end{equation}

\begin{equation}
  b_c = \mathit{high\_mult} \times E[D]
\end{equation}

\begin{equation}
  Q_j =
  \begin{cases}
    0,
      & B_j \in [a_c,\, b_c] \\[6pt]
    \dfrac{\min\!\left(|B_j - a_c|,\;|B_j - b_c|\right)}{b_c - a_c},
      & \text{otherwise}
  \end{cases}
\end{equation}

\paragraph{Step 4: Aggregate TDAS.}
For an itinerary with $M$ transport segments:

\begin{equation}
  \boxed{
    \mathrm{TDAS} = 1 - \frac{1}{M}\sum_{j=1}^{M} min(1, Q_j)
  }
\end{equation}

\paragraph{Interpretation.}
\begin{itemize}
  \item $\mathrm{TDAS} \approx 1$: Delays are highly absorbable; the itinerary
        remains sTable
  \item $\mathrm{TDAS} \approx 0$: High disruption risk due to insufficient
        buffer allocation.
\end{itemize}
\begin{table}[H]
\centering
\caption{Data entries in the UTP-Bench dataset collected from real-world sources.}
\label{tab:db_entries}
\begin{tabular}{l r}
\toprule
\textbf{Database} & \textbf{Data Entries (\#)} \\
\midrule
City Set & 20 \\
Trains & 35,117 \\
Buses & 11,313 \\
Flights & 1,198 \\
Restaurants & 244 \\
Attractions & 242 \\
Accommodations & 325 \\
Nearest Transit Stop & 861 \\
Road Delay Routes & 380 \\
\bottomrule
\end{tabular}
\end{table}

\clearpage
\section{Prompt and Annotation Details}

\begin{center}
\label{app:prompts}

\subsection{Query Generation using GPT-5}
\begin{center}
\begin{tcolorbox}[enhanced,breakable,colback=blue!5,colframe=blue!80,fonttitle=\bfseries,width=0.95\textwidth,arc=1mm,boxrule=0.5pt]
\scriptsize
You are an expert at generating natural-language travel planning queries.
Given an input Travel \{JSON\}, generate a concise and natural English query describing the trip.
In the JSON, `org' denotes the departure city. When `days' exceeds 3, `visiting\_city\_number' specifies the number of cities to be covered in the destination state (use the `target\_state' in the query). The budget is in INR. Please disregard the `level' attribute.
Here are four examples.

\smallskip
\textbf{****** Example ******}

\textbf{-----EXAMPLE 1-----}

\textbf{JSON:}

\texttt{\{"org": "Jaipur", "dest": "Nagpur", "days": 3, "visiting\_city\_number": 1, "date": ["2025-11-02", "2025-11-03", "2025-11-04"], "people\_number": 1, "local\_constraint": \{"house\_rules": null, "house\_type": null, "facilities\_amenities": null, "services": null, "cuisine": null, "transportation": null, "event": null, "attraction": null\}, "budget": 18673, "query": null, "level": "easy"\}}

\texttt{\{generated\_query\_english:"Plan a 3-day trip for 1 person from Jaipur to Nagpur from November 2nd to November 4th, 2025, with a budget of INR 18673."\}}

\textbf{-----EXAMPLE 2-----}

\textbf{JSON:}

\texttt{\{"org": "Jabalpur", "dest": ["Thiruvananthapuram", "Kochi"], "target\_state": "Kerala", "days": 5, "visiting\_city\_number": 2, "date": ["2025-11-14", "2025-11-15", "2025-11-16", "2025-11-17", "2025-11-18"], "people\_number": 4, "local\_constraint": \{"house\_rules": null, "house\_type": null, "facilities\_amenities": "Dining\_Drinks", "services": null, "cuisine": null, "transportation": null, "event": null, "attraction": null\}, "budget": 57671, "level": "medium"\}}

\texttt{\{generated\_query\_english:"Organize a 5-day itinerary for 4 people traveling from Jabalpur to explore 2 cities in Kerala, between November 14th and November 18th, 2025. The budget is INR 57671, and accommodations should include Dining and Drinks."\}}

\textbf{-----EXAMPLE 3-----}

\textbf{JSON:}

\texttt{\{"org": "Jaipur", "dest": ["Jamnagar", "Vadodara", "Ahmedabad"], "target\_state": "Gujarat", "days": 7, "visiting\_city\_number": 3, "date": ["2025-11-08", "2025-11-09", "2025-11-10", "2025-11-11", "2025-11-12", "2025-11-13", "2025-11-14"], "people\_number": 2, "local\_constraint": \{"house\_rules": "No\_Pets", "house\_type": "Luxury\_Boutique", "facilities\_amenities": null, "services": "Entertainment", "cuisine": null, "transportation": null, "event": null, "attraction": null\}, "budget": 65466, "level": "hard"\}}

\texttt{\{generated\_query\_english:"Create a detailed 7-day travel plan for 2 individuals starting from Jaipur and visiting 3 cities in Gujarat between November 8th and November 14th, 2025. The budget is INR 65466. Accommodations should be Luxury Boutique and strictly No Pets. The plan should also include Entertainment services."\}}

\textbf{****** Example Ends ******}

\bigskip
\textbf{Input:}

Travel JSON: \{JSON\}

\bigskip
\textbf{Output Requirements:}
\begin{itemize}
    \item Output must be valid JSON.
    \item The response must start with the key \texttt{"generated\_query\_english"}.
\end{itemize}

\textbf{Output format:}

\texttt{[\{"generated\_query\_english": "<generated English query>"\}]}
\end{tcolorbox}
\end{center}
\end{center}

\onecolumn
\subsection{Prompt to Generate Uncertainty Plans}

\begin{center}
\begin{tcolorbox}[enhanced,breakable,colback=blue!5,colframe=blue!80,fonttitle=\bfseries,width=0.95\textwidth,arc=1mm,boxrule=0.5pt]
\scriptsize
You are a proficient planner. Based on the provided information, query and persona, please give a detailed travel plan, including specifics such as flight numbers (e.g., 6E-789), restaurant names, and accommodation names. Note that all the information in your plans should be derived from the provided data. You must adhere to the format given in the example. Additionally, all details should align with common sense.

\medskip
\textbf{CRITICAL INSTRUCTION: INCORPORATE UNCERTAINTY \& BUFFER TIMES}

You must account for uncertainty in your planning by incorporating buffer times based on the historical data provided in the context:
\begin{enumerate}
    \item \textbf{Attractions \& Restaurants:} Use the provided \texttt{besttime\_analysis} to gauge crowd levels. If visiting during `High' or `Above average' intensity hours, add a 15--30 minute buffer to the visit duration. Try to avoid \texttt{peak\_hours}.
    \item \textbf{Road Transportation} (which include bus and Cab Transportation): In addition to \texttt{Total Travel Time}, add the \texttt{Estimated Historical Delay} to the travel duration to ensure realistic arrival times.
    \item \textbf{Train Transportation:} Check \texttt{avg\_delay\_1month} (or similar available delay metric) and add this expected delay to the train's arrival time before scheduling the next activity.
    \item \textbf{Flight Transportation:} Review \texttt{OnTimePerformance} and \texttt{DelayStats}. If the flight has a low on-time rating or high delay probability, add a 30--45 minute buffer to the arrival time before the next scheduled item.
\end{enumerate}

The symbol `\texttt{-}' indicates that information is unnecessary. For example, in the provided sample, you do not need to plan after returning to the departure city. When you travel to two cities in one day, you should note it in the \texttt{"Current City"} section as in the example (i.e., from A to B). Include events happening on that day, if any. Provide a Point of Interest List, which is an ordered list of places visited throughout the day. This list should include accommodations, attractions, or restaurants and their starting and ending timestamps. Each day must start and end with the accommodation where the traveler is staying. Breakfast is ideally scheduled at 9:40 AM and lasts about 50 minutes. Lunch is best planned for 2:20 PM, with a duration of around an hour. Dinner should take place at 8:45 PM, lasting approximately 1 hour and 15 minutes. Laidback Travelers typically explore one attraction per day and sometimes opt for more, while Adventure Seekers often visit 2 or 3 attractions, occasionally exceeding that number.

\medskip
\textbf{IMPORTANT: YOUR OUTPUT MUST BE IN JSONL FORMAT ONLY. Do not include any explanations, markdown formatting, or text outside the JSON structure.}

\medskip
\textbf{=== COMPRESSED JSON DATA FORMAT GUIDE ===}

\medskip
You are given venue crowd and timing data in a COMPRESSED JSON format to reduce context size.

\medskip
\textbf{IMPORTANT:}
\begin{itemize}
    \item This compressed format is lossless.
    \item Do NOT assume missing information.
    \item All times are in 24-hour format.
    \item Interpret numeric ranges inclusively (e.g., [11,16] = 11:00--16:59).
\end{itemize}

\textbf{GLOBAL FIELDS} (apply to the venue / average day):
\begin{itemize}
    \item \texttt{"o"}: venue opening hour
    \item \texttt{"c"}: venue closing hour
    \item \texttt{"s"}: surge behavior
    \begin{itemize}
        \item \texttt{"in"}: hour when most visitors arrive
        \item \texttt{"out"}: hour when most visitors leave
    \end{itemize}
    \item \texttt{"I"}: intensity dictionary (code $\rightarrow$ meaning)
\end{itemize}

\textbf{AVERAGE DAY OBJECT FORMAT:}

The key \texttt{"avg\_day"} represents aggregated crowd behavior across days.

Keys inside \texttt{"avg\_day"}:
\begin{itemize}
    \item \texttt{"cw"}: crowd windows, each as \texttt{[start\_hour, end\_hour, intensity\_code]}
    \item \texttt{"q"}: quiet hours (lowest crowd)
    \item \texttt{"p"}: peak window as \texttt{[start, end]} (if present)
\end{itemize}

\textbf{INTENSITY CODES:}
\begin{itemize}
    \item \texttt{BA} $\rightarrow$ Below average crowd
    \item \texttt{A} $\rightarrow$ Average crowd
    \item \texttt{AA} $\rightarrow$ Above average crowd
    \item \texttt{H} $\rightarrow$ High crowd
    \item \texttt{L} $\rightarrow$ Low crowd
\end{itemize}

\textbf{HOW TO INTERPRET:}
\begin{itemize}
    \item Use \texttt{"cw"} to reason about time-dependent crowd levels.
    \item Prefer \texttt{"q"} hours for low waiting time and higher feasibility.
    \item Treat \texttt{"p"} as the busiest continuous window.
    \item Use surge times to anticipate entry and exit congestion.
    \item Do NOT expand or rewrite the data unless explicitly asked.
\end{itemize}

\textbf{GOAL:} Use this data to perform crowd-aware reasoning, scheduling, uncertainty modeling, or feasibility checks without adding new assumptions.

\medskip
\textbf{=== END OF COMPRESSED JSON DATA FORMAT GUIDE ===}

\medskip
\textbf{****** Example ******}

\medskip
\textbf{Query:} Plan a 7-day trip for 2 people starting from Bhopal to visit 3 cities---Kanpur, Surat, and Pune---from November 6th to November 12th, 2025, with a budget of INR 91306. Accommodations should feature a Wellness Spa, provide Cleaning and Laundry services, and offer African cuisine.

\medskip
\textbf{Traveler Persona:}

Traveler Type: Culture Enthusiast; Purpose of Travel: Cultural Immersion; Spending Preference: Luxury; Location Preference: Historical Sites

\medskip
\textbf{Travel Plan:}

\begin{verbatim}
[
  {
    "days": 1,
    "current_city": "from Bhopal to Kanpur",
    "transportation": "Train Number: 12191, from Bhopal to Kanpur,
                       Departure Time: 06:30, Arrival Time: 15:15",
    "breakfast": "-",
    "attraction": "Cawnpore Club, Kanpur",
    "lunch": "-",
    "dinner": "Haveli Restaurant, Kanpur",
    "accommodation": "Hotel Bhagyaraj Palace, Kanpur",
    "event": "-",
    "point_of_interest": "Hotel Bhagyaraj Palace, check-in from 15:45
      to 16:15, nearest transit: Kanpur Central Station, 3km away;
      Cawnpore Club, visit from 16:30 to 18:30, nearest transit:
      Civil Lines, 0.5km away; Haveli Restaurant, visit from 20:45
      to 22:00, nearest transit: Mall Road Stop, 1km away; Hotel
      Bhagyaraj Palace, stay from 22:15 to 08:00, nearest transit:
      Kanpur Central Station, 3km away."
  },
  {
    "days": 2,
    "current_city": "Kanpur",
    "transportation": "-",
    "breakfast": "Little Chef Restaurant, Kanpur",
    "attraction": "Brahmavart Ghat, Kanpur",
    "lunch": "Haveli Restaurant, Kanpur",
    "dinner": "Dadi ki rasoi, Kanpur",
    "accommodation": "Hotel Bhagyaraj Palace, Kanpur",
    "event": "-",
    "point_of_interest": "Hotel Bhagyaraj Palace, stay from 08:00 to
      09:40, nearest transit: Kanpur Central Station, 3km away;
      Little Chef Restaurant, visit from 09:40 to 10:30, nearest
      transit: Mall Road Stop, 1km away; Brahmavart Ghat, visit
      from 11:00 to 13:30, nearest transit: Bithoor Ghat, 0.2km
      away; Haveli Restaurant, visit from 14:20 to 15:20, nearest
      transit: Mall Road Stop, 1km away; Dadi ki rasoi, visit from
      20:45 to 22:00, nearest transit: Bithoor Road Stop, 1.5km
      away; Hotel Bhagyaraj Palace, stay from 22:15 to 06:00,
      nearest transit: Kanpur Central Station, 3km away."
  },
  {
    "days": 3,
    "current_city": "from Kanpur to Surat",
    "transportation": "Flight Number: 6E-2145, from Kanpur to Surat,
                       Departure Time: 08:00, Arrival Time: 10:30",
    "breakfast": "-",
    "attraction": "Dutch Garden, Surat",
    "lunch": "Mysore Cafe, Surat",
    "dinner": "Levvel 5 Terrace Restro & Cafe, Surat",
    "accommodation": "Surat Marriott Hotel, Surat",
    "event": "-",
    "point_of_interest": "Hotel Bhagyaraj Palace, checkout from 06:00
      to 06:30, nearest transit: Kanpur Central Station, 3km away;
      Kanpur Airport, arrive from 07:00 to 07:30, nearest transit:
      Airport Road, 2km away; Surat Marriott Hotel, check-in from
      11:00 to 11:30, nearest transit: Surat Railway Station, 4km
      away; Dutch Garden, visit from 12:00 to 13:30, nearest
      transit: Athwalines, 1km away; Mysore Cafe, visit from 14:20
      to 15:20, nearest transit: Ring Road Stop, 1km away; Levvel
      5 Terrace Restro & Cafe, visit from 20:45 to 22:00, nearest
      transit: VIP Road Stop, 1km away; Surat Marriott Hotel, stay
      from 22:15 to 08:00, nearest transit: Surat Railway Station,
      4km away."
  },
  {
    "days": 4,
    "current_city": "Surat",
    "transportation": "-",
    "breakfast": "Taste Of India Restaurant, Surat",
    "attraction": "ISKCON Temple, Surat",
    "lunch": "Levvel 5 Terrace Restro & Cafe, Surat",
    "dinner": "Mysore Cafe, Surat",
    "accommodation": "Surat Marriott Hotel, Surat",
    "event": "-",
    "point_of_interest": "Surat Marriott Hotel, stay from 08:00 to
      09:40, nearest transit: Surat Railway Station, 4km away;
      Taste Of India Restaurant, visit from 09:40 to 10:30, nearest
      transit: Varachha Road, 1km away; ISKCON Temple, visit from
      11:00 to 13:30, nearest transit: Athwalines, 0.5km away;
      Levvel 5 Terrace Restro & Cafe, visit from 14:20 to 15:20,
      nearest transit: VIP Road Stop, 1km away; Mysore Cafe, visit
      from 20:45 to 22:00, nearest transit: Ring Road Stop, 1km
      away; Surat Marriott Hotel, stay from 22:15 to 07:00, nearest
      transit: Surat Railway Station, 4km away."
  },
  {
    "days": 5,
    "current_city": "from Surat to Pune",
    "transportation": "Flight Number: 6E-3456, from Surat to Pune,
                       Departure Time: 09:00, Arrival Time: 10:30",
    "breakfast": "-",
    "attraction": "Mahadaji Shinde Chatri, Pune",
    "lunch": "Spice Kitchen, Pune",
    "dinner": "Sante Spa Cuisine, Pune",
    "accommodation": "Conrad Pune, Pune",
    "event": "-",
    "point_of_interest": "Surat Marriott Hotel, checkout from 07:00
      to 07:30, nearest transit: Surat Railway Station, 4km away;
      Surat Airport, arrive from 08:00 to 08:30, nearest transit:
      Airport Road, 3km away; Conrad Pune, check-in from 11:00 to
      11:30, nearest transit: Pune Junction, 5km away; Mahadaji
      Shinde Chatri, visit from 12:00 to 13:30, nearest transit:
      Wanowrie, 1km away; Spice Kitchen, visit from 14:20 to 15:20,
      nearest transit: Koregaon Park, 0.5km away; Sante Spa Cuisine,
      visit from 20:45 to 22:00, nearest transit: Koregaon Park,
      0.3km away; Conrad Pune, stay from 22:15 to 08:00, nearest
      transit: Pune Junction, 5km away."
  },
  {
    "days": 6,
    "current_city": "Pune",
    "transportation": "-",
    "breakfast": "Spice Kitchen, Pune",
    "attraction": "ISKCON NVCC Temple, Pune",
    "lunch": "Sante Spa Cuisine, Pune",
    "dinner": "The Flour Works, Pune",
    "accommodation": "Conrad Pune, Pune",
    "event": "-",
    "point_of_interest": "Conrad Pune, stay from 08:00 to 09:40,
      nearest transit: Pune Junction, 5km away; Spice Kitchen,
      visit from 09:40 to 10:30, nearest transit: Koregaon Park,
      0.5km away; ISKCON NVCC Temple, visit from 11:00 to 13:30,
      nearest transit: NIBM Road, 1km away; Sante Spa Cuisine,
      visit from 14:20 to 15:20, nearest transit: Koregaon Park,
      0.3km away; The Flour Works, visit from 20:45 to 22:00,
      nearest transit: Kalyani Nagar, 1km away; Conrad Pune, stay
      from 22:15 to 07:00, nearest transit: Pune Junction, 5km
      away."
  },
  {
    "days": 7,
    "current_city": "from Pune to Bhopal",
    "transportation": "Flight Number: 6E-5432, from Pune to Bhopal,
                       Departure Time: 11:00, Arrival Time: 12:30",
    "breakfast": "Kumar Pacific Mall, Pune",
    "attraction": "-",
    "lunch": "-",
    "dinner": "-",
    "accommodation": "-",
    "event": "-",
    "point_of_interest": "Conrad Pune, checkout from 07:00 to 07:30,
      nearest transit: Pune Junction, 5km away; Kumar Pacific Mall,
      visit from 07:45 to 08:45, nearest transit: Swargate, 1km
      away; Pune Airport, arrive from 09:30 to 10:30, nearest
      transit: Airport Road, 2km away; Flight 6E-5432, travel from
      11:00 to 12:30."
  }
]
\end{verbatim}

\textbf{****** Example Ends ******}

\medskip
\textbf{Given information:} \{enriched\_ref\_info\}

\textbf{Query:} \{query\}

\textbf{Traveler Persona:}

\{persona\}

\medskip
\textbf{Output (JSONL):}

\end{tcolorbox}
\end{center}

\onecolumn
\begin{center}
\subsection{Prompt to fill the realistic timings}

\begin{tcolorbox}[enhanced,breakable,colback=blue!5,colframe=blue!80,fonttitle=\bfseries,width=\textwidth,arc=1mm,boxrule=0.5pt]
\scriptsize

You are a proficient travel plan time allocator. You do \textbf{NOT} design new itineraries from scratch.

\smallskip
Your \textbf{ONLY} job is to take an \textbf{ALREADY GENERATED TRAVEL PLAN} (with days, cities, attractions, restaurants, and accommodations already chosen) that is missing some or all timing information, and \textbf{FILL IN REALISTIC, CONSISTENT TIMINGS} for every activity, stay, visit, and transport segment.

\smallskip
You must:
\begin{itemize}
    \item Preserve the existing structure of the plan (days, current\_city, transportation, breakfast, attraction, lunch, dinner, accommodation, event, and the text segments inside point\_of\_interest / point\_of\_interest\_list).
    \item Preserve the exact list and order of cities, accommodations, attractions, restaurants, events, and transit segments.
    \item ONLY add or adjust times and clearly stated buffers; do NOT add new places or remove existing ones.
    \item Ensure that the final plan is temporally consistent (no overlaps, no negative gaps, realistic travel + activity durations) and respects uncertainty and buffer rules.
\end{itemize}

\smallskip
\textbf{CRITICAL INPUT FORMAT}

\smallskip
You receive each sample as a JSON object with the following structure (in JSONL format, one object per line). The field \texttt{JSON} contains information like: origin city, destination city (or list of cities), dates array (one date per day), number of days, number of people, traveler type/persona, local constraints, budget, and the natural language query. The field \texttt{plan} is an array of per-day objects where transportation, meals, attraction, accommodation, and event are already chosen and \textbf{MUST NOT} be changed. The \texttt{point\_of\_interest} or \texttt{point\_of\_interest\_list} is a semicolon-separated ordered list of segments describing stays, check-in/checkout windows, visits, and travel segments. In many inputs, some time windows are missing and represented as empty parentheses: \texttt{()}.

\smallskip
Your task is to \textbf{REPLACE ALL SUCH EMPTY PARENTHESIS TIME SLOTS WITH CONCRETE 24-HOUR TIMES (HH:MM)}, while keeping the rest of the text unchanged.

\smallskip
\textbf{IMPORTANT:} You may slightly adjust already filled times (e.g., shift by 10--30 minutes) \textit{only} if necessary to maintain chronological consistency, to account for uncertainty buffers, or to avoid overlaps. Otherwise, keep existing times as-is.

\smallskip
\textbf{UNCERTAINTY \& BUFFER RULES (MUST FOLLOW)}

\begin{enumerate}
% [noitemsep,topsep=2pt,parsep=0pt,partopsep=0pt]
    \item \textbf{Attractions \& Restaurants (crowd-aware):} When \texttt{besttime\_analysis} or similar crowd data is available, use its crowd intensity windows. If an attraction/restaurant is visited during a ``High'' or ``Above average'' intensity window, add a 15--30 minute buffer to the base visit duration. Prefer visits during quieter windows when possible, but do NOT change the chosen venue or day. Avoid \texttt{peak\_hours} where explicitly marked, or if unavoidable, add an appropriate wait-time buffer.
    \item \textbf{Road Transportation (buses, cabs, cars):} When total travel time and Estimated Historical Delay are provided, the scheduled travel window must be: base\_travel\_duration + estimated\_delay.
    \item \textbf{Train Transportation:} Use available delay metrics such as \texttt{avg\_delay\_1month} to extend the arrival time. The planned arrival time must be: scheduled\_arrival + expected\_delay. Ensure subsequent activities start AFTER this buffered arrival time.
    \item \textbf{Flight Transportation:} Use \texttt{OnTimePerformance} and \texttt{DelayStats} where provided. If the flight has a poor on-time record or high delay probability, add a 30--45 minute buffer after the scheduled arrival before starting the next activity.
\end{enumerate}

\smallskip
\textbf{DAY STRUCTURE \& MEAL-TIME HEURISTICS}

\smallskip
When filling times, follow these heuristics unless they clearly conflict with already-fixed times or explicit constraints:
\begin{itemize}
% [noitemsep,topsep=2pt,parsep=0pt,partopsep=0pt]
    \item Each day should start and end at the accommodation (except pure travel days with final return to origin).
    \item Morning wake-up / stay at accommodation: around 07:00--09:40.
    \item Breakfast: start around 09:40, duration $\sim$50 minutes.
    \item Late-morning attractions: roughly 10:30--13:30.
    \item Lunch: start around 14:20, duration $\sim$60 minutes.
    \item Afternoon attractions/activities: roughly 15:00--18:30.
    \item Evening free time / events: roughly 18:30--20:30.
    \item Dinner: start around 20:45, duration $\sim$75 minutes.
    \item Night stay at accommodation: from around 22:15 until next morning 06:00--08:00.
\end{itemize}

\smallskip
\textbf{TRAVELER PERSONA \& ATTRACTION DENSITY}

\smallskip
You must respect the intended intensity of sightseeing implied by the traveler type:
\begin{itemize}
% [noitemsep,topsep=2pt,parsep=0pt,partopsep=0pt]
    \item \textbf{Risk Tolerant:} Strict timings, shorter buffers, close to or below ideal buffer duration. Typically 2--3 attractions per day, sometimes more.
    \item \textbf{Risk Optimized:} Balanced schedule, moderate buffer equal to or slightly more than ideal buffer duration.
    \item \textbf{Risk Averse:} Significantly larger buffers,may go upto twice or well beyond ideal buffer duration; may prefer earlier returns to accommodation.
\end{itemize}

\smallskip
Do NOT change which attractions are visited. Use these rules only to set visit durations and buffers.

\smallskip
\textbf{TEMPORAL CONSISTENCY RULES}
\begin{itemize}
% [noitemsep,topsep=2pt,parsep=0pt,partopsep=0pt]
    \item All segments in \texttt{point of interest} for each day must be chronological (each start time $\geq$ previous end time, with realistic small transfer gaps).
    \item Transport windows must be long enough to cover base travel time + uncertainty buffers.
    \item Check-in/check-out windows must surround realistic arrival/departure times.
    \item Never schedule an attraction or restaurant before the traveler physically arrives in that city.
    \item The final stay at accommodation must reach at least 06:00--08:00 the next morning, except for overnight transport.
\end{itemize}

\smallskip
\textbf{OUTPUT REQUIREMENTS (VERY IMPORTANT)}
\begin{itemize}
% [noitemsep,topsep=2pt,parsep=0pt,partopsep=0pt]
    \item Output \textbf{MUST} be in valid JSONL format ONLY each line is one JSON object with the same schema as the input.
    \item For each input object, output exactly one JSON object with the same \texttt{id} and \texttt{JSON} fields.
    \item Preserve all non-time textual content (city names, attraction names, restaurant names, accommodation names, nearest transit info, delay descriptions, etc.).
    \item Replace every occurrence of \texttt{()} with concrete 24-hour times in \texttt{HH:MM} format.
    \item Ensure every segment ends up with a well-defined start and end time, or explicit \texttt{Departure Time} and \texttt{Arrival Time} fields filled.
    \item Do NOT output any explanations, comments, or markdown formatting outside of the JSON objects.
\end{itemize}

\smallskip
\textbf{Given information:} \{enriched\_ref\_info\}

\smallskip
\textbf{Input (JSONL with existing plan and missing times):} \{input\_plan\_with\_missing\_times\}, reference\_information

\smallskip
\textbf{Output (JSONL with same structure and all times filled)}

\end{tcolorbox}
\end{center}

\onecolumn
\subsection{Prompt to Generate Without-Uncertainty Plans}
% \begin{document}
 
\begin{center}
\begin{tcolorbox}[enhanced,breakable,colback=blue!5,colframe=blue!80,fonttitle=\bfseries,width=0.95\textwidth,arc=1mm,boxrule=0.5pt]
\scriptsize
You are a proficient planner. Based on the provided information, query and persona, please give a detailed travel plan, including specifics such as flight numbers (e.g., 6E-789), restaurant names, and accommodation names. Note that all the information in your plans should be derived from the provided data. You must adhere to the format given in the example. Additionally, all details should align with common sense.
 
\medskip
The symbol `\texttt{-}' indicates that information is unnecessary. For example, in the provided sample, you do not need to plan after returning to the departure city. When you travel to two cities in one day, you should note it in the \texttt{"Current City"} section as in the example (i.e., from A to B). Include events happening on that day, if any. Provide a Point of Interest List, which is an ordered list of places visited throughout the day. This list should include accommodations, attractions, or restaurants and their starting and ending timestamps. Each day must start and end with the accommodation where the traveler is staying.
 
\medskip
\textbf{****** Example ******}
 
\medskip
\textbf{Query:} Plan a 7-day trip for 2 people starting from Bhopal to visit 3 cities---Kanpur, Surat, and Pune---from November 6th to November 12th, 2025, with a budget of INR 91306. Accommodations should feature a Wellness Spa, provide Cleaning and Laundry services, and offer African cuisine.
 
\medskip
\textbf{Traveler Persona:}
 
Traveler Type: Culture Enthusiast; Purpose of Travel: Cultural Immersion; Spending Preference: Luxury; Location Preference: Historical Sites
 
\medskip
\textbf{Travel Plan:}
 
\begin{verbatim}
{
  "plan": [
    {
      "day": 1,
      "current_city": "from Bhopal to Kanpur",
      "transportation": "Train Number: 12191, from Bhopal to Kanpur,
                         Departure Time: 06:30, Arrival Time: 15:15",
      "breakfast": "-",
      "attraction": "Cawnpore Club, Kanpur",
      "lunch": "-",
      "dinner": "Haveli Restaurant, Kanpur",
      "accommodation": "Hotel Bhagyaraj Palace, Kanpur",
      "event": "-",
      "point_of_interest_list": "Hotel Bhagyaraj Palace, check-in from
        15:45 to 16:15, nearest transit: Kanpur Central Station, 11km
        away; Cawnpore Club, visit from 16:30 to 18:30, nearest transit:
        Civil Lines, 0.5km away; Haveli Restaurant, visit from 20:45 to
        22:00, nearest transit: Mall Road Stop, 1km away; Hotel Bhagyaraj
        Palace, stay from 22:15 to 08:00, nearest transit: Kanpur Central
        Station, 3km away."
    },
    {
      "day": 2,
      "current_city": "Kanpur",
      "transportation": "-",
      "breakfast": "Little Chef Restaurant, Kanpur",
      "attraction": "Brahmavart Ghat, Kanpur",
      "lunch": "Haveli Restaurant, Kanpur",
      "dinner": "Dadi ki rasoi, Kanpur",
      "accommodation": "Hotel Bhagyaraj Palace, Kanpur",
      "event": "-",
      "point_of_interest_list": "Hotel Bhagyaraj Palace, stay from 08:00
        to 09:40, nearest transit: Kanpur Central Station, 3km away;
        Little Chef Restaurant, visit from 09:40 to 10:30, nearest
        transit: Mall Road Stop, 1km away; Brahmavart Ghat, visit from
        11:00 to 13:30, nearest transit: Bithoor Ghat, 0.2km away;
        Haveli Restaurant, visit from 14:20 to 15:20, nearest transit:
        Mall Road Stop, 1km away; Dadi ki rasoi, visit from 20:45 to
        22:00, nearest transit: Bithoor Road Stop, 6km away; Hotel
        Bhagyaraj Palace, stay from 22:15 to 06:00, nearest transit:
        Kanpur Central Station, 3km away."
    },
    {
      "day": 3,
      "current_city": "from Kanpur to Surat",
      "transportation": "Flight Number: 6E-2145, from Kanpur to Surat,
                         Departure Time: 08:00, Arrival Time: 10:30",
      "breakfast": "-",
      "attraction": "Dutch Garden, Surat",
      "lunch": "Mysore Cafe, Surat",
      "dinner": "Levvel 5 Terrace Restro & Cafe, Surat",
      "accommodation": "Surat Marriott Hotel, Surat",
      "event": "-",
      "point_of_interest_list": "Hotel Bhagyaraj Palace, checkout from
        06:00 to 06:30, nearest transit: Kanpur Central Station, 3km
        away; Kanpur Airport, arrive from 07:00 to 07:30, nearest
        transit: Airport Road, 2km away; Surat Marriott Hotel, check-in
        from 11:00 to 11:30, nearest transit: Surat Railway Station,
        4km away; Dutch Garden, visit from 12:00 to 13:30, nearest
        transit: Athwalines, 1km away; Mysore Cafe, visit from 14:20
        to 15:20, nearest transit: Ring Road Stop, 1km away; Levvel 5
        Terrace Restro & Cafe, visit from 20:45 to 22:00, nearest
        transit: VIP Road Stop, 1km away; Surat Marriott Hotel, stay
        from 22:15 to 08:00, nearest transit: Surat Railway Station,
        4km away."
    },
    {
      "day": 4,
      "current_city": "Surat",
      "transportation": "-",
      "breakfast": "Taste Of India Restaurant, Surat",
      "attraction": "ISKCON Temple, Surat",
      "lunch": "Levvel 5 Terrace Restro & Cafe, Surat",
      "dinner": "Mysore Cafe, Surat",
      "accommodation": "Surat Marriott Hotel, Surat",
      "event": "-",
      "point_of_interest_list": "Surat Marriott Hotel, stay from 08:00
        to 09:40, nearest transit: Surat Railway Station, 4km away;
        Taste Of India Restaurant, visit from 09:40 to 10:30, nearest
        transit: Varachha Road, 1km away; ISKCON Temple, visit from
        11:00 to 13:30, nearest transit: Athwalines, 0.5km away;
        Levvel 5 Terrace Restro & Cafe, visit from 14:20 to 15:20,
        nearest transit: VIP Road Stop, 1km away; Mysore Cafe, visit
        from 20:45 to 22:00, nearest transit: Ring Road Stop, 1km
        away; Surat Marriott Hotel, stay from 22:15 to 07:00, nearest
        transit: Surat Railway Station, 4km away."
    },
    {
      "day": 5,
      "current_city": "from Surat to Pune",
      "transportation": "Flight Number: 6E-3456, from Surat to Pune,
                         Departure Time: 09:00, Arrival Time: 10:30",
      "breakfast": "-",
      "attraction": "Mahadaji Shinde Chatri, Pune",
      "lunch": "Spice Kitchen, Pune",
      "dinner": "Sante Spa Cuisine, Pune",
      "accommodation": "Conrad Pune, Pune",
      "event": "-",
      "point_of_interest_list": "Surat Marriott Hotel, checkout from
        07:00 to 07:30, nearest transit: Surat Railway Station, 4km
        away; Surat Airport, arrive from 08:00 to 08:30, nearest
        transit: Airport Road, 3km away; Conrad Pune, check-in from
        11:00 to 11:30, nearest transit: Pune Junction, 5km away;
        Mahadaji Shinde Chatri, visit from 12:00 to 13:30, nearest
        transit: Wanowrie, 1km away; Spice Kitchen, visit from 14:20
        to 15:20, nearest transit: Koregaon Park, 0.5km away; Sante
        Spa Cuisine, visit from 20:45 to 22:00, nearest transit:
        Koregaon Park, 0.3km away; Conrad Pune, stay from 22:15 to
        08:00, nearest transit: Pune Junction, 5km away."
    },
    {
      "day": 6,
      "current_city": "Pune",
      "transportation": "-",
      "breakfast": "Spice Kitchen, Pune",
      "attraction": "ISKCON NVCC Temple, Pune",
      "lunch": "Sante Spa Cuisine, Pune",
      "dinner": "The Flour Works, Pune",
      "accommodation": "Conrad Pune, Pune",
      "event": "-",
      "point_of_interest_list": "Conrad Pune, stay from 08:00 to 09:40,
        nearest transit: Pune Junction, 5km away; Spice Kitchen, visit
        from 09:40 to 10:30, nearest transit: Koregaon Park, 0.5km
        away; ISKCON NVCC Temple, visit from 11:00 to 13:30, nearest
        transit: NIBM Road, 1km away; Sante Spa Cuisine, visit from
        14:20 to 15:20, nearest transit: Koregaon Park, 0.3km away;
        The Flour Works, visit from 20:45 to 22:00, nearest transit:
        Kalyani Nagar, 1km away; Conrad Pune, stay from 22:15 to
        07:00, nearest transit: Pune Junction, 5km away."
    },
    {
      "day": 7,
      "current_city": "from Pune to Bhopal",
      "transportation": "Flight Number: 6E-5432, from Pune to Bhopal,
                         Departure Time: 11:00, Arrival Time: 12:30",
      "breakfast": "Kumar Pacific Mall, Pune",
      "attraction": "-",
      "lunch": "-",
      "dinner": "-",
      "accommodation": "-",
      "event": "-",
      "point_of_interest_list": "Conrad Pune, checkout from 07:00 to
        07:30, nearest transit: Pune Junction, 5km away; Kumar Pacific
        Mall, visit from 07:45 to 08:45, nearest transit: Swargate,
        1km away; Pune Airport, arrive from 09:30 to 10:30, nearest
        transit: Airport Road, 2km away; Flight 6E-5432, travel from
        11:00 to 12:30."
    }
  ]
}
\end{verbatim}
 
\textbf{****** Example Ends ******}
 
\medskip
\textbf{Given information:} \{enriched\_ref\_info\}
 
\textbf{Query:} \{query\}
 
\textbf{Traveler Persona:}
 
\{persona\}
 
\medskip
\textbf{Output:}
 
\end{tcolorbox}
\end{center}
\section{Case Studies}
\label{app:case-studies}

\begin{figure}[h]
\centering
\begin{tcolorbox}[enhanced,
    colback=yellow!15,
    colframe=black,
    boxrule=0.5pt,
    arc=1mm,
    width=0.45\textwidth,
    fontupper=\scriptsize]
\textbf{Query:} Plan a 3-day trip for 4 people from Bhopal
to Kanpur from November 19th to November 21st, 2025, with
a budget of INR 123211. Traveler Type: Laidback Traveler;
Risk Averse; Spending: Luxury.
\noindent\rule{\linewidth}{0.4pt}
\textbf{Plan:}
\texttt{\{}\\
\texttt{"days": 2,}\\
\texttt{"current\_city": "Kanpur",}\\
\texttt{"transportation": "-",}\\
\texttt{"breakfast": "Little Chef Restaurant, Kanpur",}\\
\texttt{"attraction": "Nana Rao Smarak Park, Kanpur",}\\
\texttt{"lunch": "Dadi ki Rasoi, Kanpur",}\\
\texttt{"dinner": "Mocha Cafe \& Bar, Kanpur",}\\
\texttt{"accommodation": "The Landmark Towers, Kanpur",}\\
\texttt{"point\_of\_interest": "The Landmark Towers,}\\
\texttt{\hspace{1em}stay from 08:00 to 09:40;}\\
\texttt{\hspace{1em}Little Chef Restaurant, visit from}\\
\texttt{\hspace{1em}09:55 to 10:45;}\\
\colorbox{yellow}{\texttt{\hspace{1em}Nana Rao Smarak Park, visit from}}\\
\colorbox{yellow}{\texttt{\hspace{1em}11:10 to 13:40 (150 min, generous}}\\
\colorbox{yellow}{\texttt{\hspace{1em}buffer within Risk Averse range);}}\\
\texttt{\hspace{1em}Dadi ki Rasoi, visit from 14:20}\\
\texttt{\hspace{1em}to 15:20;}\\
\colorbox{yellow}{\texttt{\hspace{1em}The Landmark Towers, rest from}}\\
\colorbox{yellow}{\texttt{\hspace{1em}15:45 to 18:50 (3.5 hr recovery}}\\
\colorbox{yellow}{\texttt{\hspace{1em}buffer, Risk Averse profile);}}\\
\texttt{\hspace{1em}Mocha Cafe \& Bar, visit from}\\
\texttt{\hspace{1em}19:05 to 20:20;}\\
\texttt{\hspace{1em}The Landmark Towers, stay from}\\
\texttt{\hspace{1em}20:40 to 07:30."}\\
\texttt{\}}
\noindent\rule{\linewidth}{0.4pt}
\textbf{Remark:} Good \textbf{BAS} score --- the planner
allocates buffer times that are well-calibrated to the
Risk Averse profile, which demands larger recovery margins
at every POI. Nana Rao Smarak Park receives a generous
150-min visit window, and a 3.5-hour rest block is
explicitly inserted between lunch and dinner at the hotel.
Across all three days, transitions between activities are
unhurried and visit durations consistently exceed the base
duration by a margin appropriate for a laidback, Risk Averse
traveller. The buffer at each POI is neither too tight nor
excessively conservative, sitting comfortably within the
expected range and resulting in near-zero deviation
penalties throughout.
\end{tcolorbox}
\caption{Case: Good BAS --- Risk Averse profile with well-calibrated POI buffers.}
\label{fig:case_good_bas}
\end{figure}

\begin{figure}[h]
\centering
\begin{tcolorbox}[enhanced,
    colback=yellow!15,
    colframe=black,
    boxrule=0.5pt,
    arc=1mm,
    width=0.45\textwidth,
    fontupper=\scriptsize]
\textbf{Query:} Plan a 3 days trip from Bhopal to Jaipur
from 2025-11-19 to 2025-11-21, with a budget of INR 11785.
Traveler Type: Laidback Traveler; Risk Averse; Nature lover.
\noindent\rule{\linewidth}{0.4pt}
\textbf{Plan:}
\texttt{\{}\\
\texttt{"days": 3,}\\
\texttt{"current\_city": "from Jaipur to Bhopal",}\\
\texttt{"transportation": "Train 19711 JP BPL EXP,}\\
\texttt{\hspace{1em}Jaipur to Bhopal, Dep: 18:25,}\\
\texttt{\hspace{1em}Arr: 06:45 (+1 day, includes}\\
\texttt{\hspace{1em}avg\_delay\_1month of 4 min)",}\\
\texttt{"attraction": "Sisodia Rani Palace, Jaipur",}\\
\texttt{"lunch": "Govindam Retreat, Jaipur",}\\
\texttt{"accommodation": "-",}\\
\texttt{"point\_of\_interest": "Trident Jaipur,}\\
\texttt{\hspace{1em}stay from 07:00 to 09:20;}\\
\texttt{\hspace{1em}Monarch Restaurant, visit from}\\
\texttt{\hspace{1em}09:40 to 10:30;}\\
\colorbox{yellow}{\texttt{\hspace{1em}Sisodia Rani Palace, visit from}}\\
\colorbox{yellow}{\texttt{\hspace{1em}11:00 to 14:15 (15 min buffer for}}\\
\colorbox{yellow}{\texttt{\hspace{1em}above avg crowd in late hours);}}\\
\texttt{\hspace{1em}Govindam Retreat, visit from}\\
\texttt{\hspace{1em}14:45 to 15:45;}\\
\colorbox{yellow}{\texttt{\hspace{1em}Jaipur Railway Station, arrive}}\\
\colorbox{yellow}{\texttt{\hspace{1em}from 17:40 to 17:55 (45 min pre-}}\\
\colorbox{yellow}{\texttt{\hspace{1em}departure buffer absorbing delay);}}\\
\colorbox{yellow}{\texttt{\hspace{1em}Board Train 19711 from 18:10}}\\
\colorbox{yellow}{\texttt{\hspace{1em}to 18:25 (+1 day, avg\_delay\_1month}}\\
\colorbox{yellow}{\texttt{\hspace{1em}of 4 min incorporated).}}\\
\texttt{\}}
\noindent\rule{\linewidth}{0.4pt}
\textbf{Remark:} Good \textbf{TDAS} score --- the planner
draws on historical delay data (\texttt{avg\_delay\_1month})
for both train segments and explicitly incorporates the
expected delay into the planned arrival time rather than
assuming on-time performance. A 45-minute pre-departure
buffer is built in at Jaipur Railway Station, giving the
itinerary sufficient headroom to absorb realistic train
delays without disrupting subsequent activities. This
conservative but not excessive buffer is precisely aligned
with what a Risk Averse traveller requires, ensuring the
transport segment neither runs too tight nor wastes time,
and resulting in near-zero penalty for both journeys.
\end{tcolorbox}
\caption{Case: Good TDAS --- delay-aware train buffer within Risk Averse acceptable range.}
\label{fig:case_good_tdas}
\end{figure}

\begin{figure}[h]
\centering
\begin{tcolorbox}[enhanced,
    colback=yellow!15,
    colframe=black,
    boxrule=0.5pt,
    arc=1mm,
    width=0.45\textwidth,
    fontupper=\scriptsize]

\textbf{Query:} Plan a 3 days trip for 3 people from Visakhapatnam
to Hyderabad from 2025-11-20 to 2025-11-22, with a budget of
INR 38730.

\noindent\rule{\linewidth}{0.4pt}

\textbf{Plan:}

\texttt{\{}\\
\texttt{"days": 3,}\\
\texttt{"current\_city": "from Hyderabad to Visakhapatnam",}\\
\texttt{"transportation": "Flight Number: IX-2819, from Hyderabad}\\
\texttt{\hspace{1em}to Visakhapatnam, Departure Time: 12:45,}\\
\texttt{\hspace{1em}Arrival Time: 14:55",}\\
\texttt{"breakfast": "The Park Hyderabad, Hyderabad",}\\
\texttt{"attraction": "Lumbini Park, Hyderabad",}\\
\texttt{"lunch": "-",}\\
\texttt{"dinner": "-",}\\
\texttt{"accommodation": "-",}\\
\texttt{"event": "-",}\\
\texttt{"point\_of\_interest": "Taj Deccan, checkout from 07:00}\\
\texttt{\hspace{1em}to 07:30, nearest transit: City Center,}\\
\texttt{\hspace{1em}0.294km away; The Park Hyderabad, visit}\\
\texttt{\hspace{1em}from 07:50 to 08:50, nearest transit:}\\
\texttt{\hspace{1em}Necklace Road, 0.113km away;}\\
\texttt{\hspace{1em}}\colorbox{yellow}{\texttt{Lumbini Park, visit from () to (),}}\\
\texttt{\hspace{1em}}\colorbox{yellow}{\texttt{nearest transit: Secretariat Bus Stop,}}\\
\texttt{\hspace{1em}}\colorbox{yellow}{\texttt{0.154km away;}}\\
\texttt{\hspace{1em}Hyderabad Airport, arrive from 11:40}\\
\texttt{\hspace{1em}to 12:10; Flight IX-2819, travel from}\\
\texttt{\hspace{1em}12:45 to 14:55 (added 40 min buffer}\\
\texttt{\hspace{1em}for possible delay)."}\\
\texttt{\}}

\noindent\rule{\linewidth}{0.4pt}

\textbf{Remark:} In this given example, the language agent
didn't fill the timings, which caused the lack of information.

\end{tcolorbox}
\caption{Case: Incosistency in filling timings.}
\label{fig:case_example}
\end{figure}

\begin{figure}[h]
\centering
\begin{tcolorbox}[enhanced,
    colback=yellow!15,
    colframe=black,
    boxrule=0.5pt,
    arc=1mm,
    width=0.45\textwidth,
    fontupper=\scriptsize]

\textbf{Query:} Plan a 3 days trip for one person from Pune to
Hyderabad from 2025-11-17 to 2025-11-19, with a budget of
INR 15743.

\noindent\rule{\linewidth}{0.4pt}

\textbf{Plan:}

\texttt{\{}\\
\texttt{"days": 1,}\\
\texttt{"current\_city": "from Pune to Hyderabad",}\\
\colorbox{yellow}{\texttt{"transportation":"Flight Number: IX-2837,}}\\
\colorbox{yellow}{\texttt{from Pune to Hyderabad, Departure Time:}}\\
\colorbox{yellow}{\texttt{04:55, Arrival Time: 06:15",}}\\
\texttt{"breakfast": "-",}\\
\texttt{"attraction": "Hussain Sagar Lake, Hyderabad",}\\
\texttt{"lunch": "Pakka Local, Hyderabad",}\\
\texttt{"dinner": "Exotica Banjara Hills- North Indian}\\
\texttt{\hspace{1em}Rooftop Restaurant, Hyderabad",}\\
\texttt{"accommodation": "The Park Hyderabad, Hyderabad",}\\
\texttt{"event": "-",}\\
\texttt{"point\_of\_interest": "Pune Airport, arrive from}\\
\texttt{\hspace{1em}04:05 to 04:25, nearest transit: Airport}\\
\texttt{\hspace{1em}Road, 2km away; Flight IX-2837, travel}\\
\texttt{\hspace{1em}from 04:55 to 06:15; The Park Hyderabad,}\\
\texttt{\hspace{1em}check-in from 06:20 to 07:40, nearest}\\
\texttt{\hspace{1em}transit: Necklace Road, 0.113km away;}\\
\texttt{\hspace{1em}Hussain Sagar Lake, visit from 10:30}\\
\texttt{\hspace{1em}to 15:00, nearest transit: Boats to}\\
\texttt{\hspace{1em}Budha Statue, 0.509km away;}\\
\colorbox{yellow}{\texttt{\hspace{1em}Pakka Local, visit from 15:30 to 16:30,}}\\
\colorbox{yellow}{\texttt{\hspace{1em}nearest transit: Kacheguda,0.317km away;}}\\
\texttt{\hspace{1em}Exotica Banjara Hills- North Indian}\\
\texttt{\hspace{1em}Rooftop Restaurant, visit from 20:00}\\
\texttt{\hspace{1em}to 21:15, nearest transit: Kaman,}\\
\texttt{\hspace{1em}0.315km away; The Park Hyderabad,}\\
\texttt{\hspace{1em}stay from 21:30 to 07:30, nearest}\\
\texttt{\hspace{1em}transit: Necklace Road, 0.113km away."}\\
\texttt{\}}

\noindent\rule{\linewidth}{0.4pt}

\textbf{Remark:} Flight \texttt{IX-2837} from Pune to Hyderabad
did not add any buffer time despite an average delay of 34 min
(based on on-time performance reference). No buffer was added
for lunch duration. Crowd information was not considered.

\end{tcolorbox}
\caption{Case: Missing buffer times and crowd information.}
\label{fig:case_example2}
\end{figure}

\begin{figure}[h]
\centering
\begin{tcolorbox}[enhanced,
    colback=yellow!15,
    colframe=black,
    boxrule=0.5pt,
    arc=1mm,
    width=0.45\textwidth,
    fontupper=\scriptsize]

\textbf{Query:} Plan a 3 days trip for 2 people from Lucknow
to Vadodara from 2025-11-25 to 2025-11-27, with a budget of
INR 15555 and want accommodation which allows pets.

\noindent\rule{\linewidth}{0.4pt}

\textbf{Plan:}

\texttt{\{}\\
\texttt{"days": 2,}\\
\texttt{"current\_city": "Vadodara",}\\
\texttt{"transportation": "-",}\\
\texttt{"breakfast": "22nd Parallel, Vadodara",}\\
\texttt{"attraction": "Maharaja Fatesingh Museum, Vadodara",}\\
\texttt{"lunch": "Candle Light Date N Dinner \&}\\
\texttt{\hspace{1em}Rooftop Restaurant, Vadodara",}\\
\texttt{"dinner": "Le Prive World Cuisine \& Pizzeria,}\\
\texttt{\hspace{1em}Vadodara",}\\
\texttt{"accommodation": "Courtyard by Marriott Vadodara,}\\
\texttt{\hspace{1em}Vadodara",}\\
\texttt{"event": "-",}\\
\texttt{"point\_of\_interest": "Courtyard by Marriott}\\
\texttt{\hspace{1em}Vadodara, stay from 07:30 to 09:10,}\\
\texttt{\hspace{1em}nearest transit: Vadodara 'C' Cabin,}\\
\texttt{\hspace{1em}0.929km away; 22nd Parallel, visit}\\
\texttt{\hspace{1em}from 09:20 to 10:10, nearest transit:}\\
\texttt{\hspace{1em}Vadodara Junction, 1.269km away;}\\
\colorbox{yellow}{\texttt{\hspace{1em}Maharaja Fatesingh Museum, visit from}}\\
\colorbox{yellow}{\texttt{\hspace{1em}10:30 to 16:30 (includes 2-hour buffer}}\\
\colorbox{yellow}{\texttt{\hspace{1em}for Above Average crowd), nearest}}\\
\colorbox{yellow}{\texttt{\hspace{1em}transit: Vishvamitri Jn. (NG), 1.369km;}}\\
\texttt{\hspace{1em}Candle Light Date N Dinner \& Rooftop}\\
\texttt{\hspace{1em}Restaurant, visit from 16:50 to 17:40,}\\
\texttt{\hspace{1em}nearest transit: Vadodara Junction,}\\
\texttt{\hspace{1em}1.111km away; Le Prive World Cuisine}\\
\texttt{\hspace{1em}\& Pizzeria, visit from 20:00 to 21:15,}\\
\texttt{\hspace{1em}nearest transit: Vadodara 'C' Cabin,}\\
\texttt{\hspace{1em}0.769km away; Courtyard by Marriott}\\
\texttt{\hspace{1em}Vadodara, stay from 21:30 to 07:00,}\\
\texttt{\hspace{1em}nearest transit: Vadodara 'C' Cabin,}\\
\texttt{\hspace{1em}0.929km away."}\\
\texttt{\}}

\noindent\rule{\linewidth}{0.4pt}

\textbf{Remark:} Unrealistically large buffer added --- a 2-hour
delay margin for an ``above average crowd'' scenario excessively
inflates the museum visit duration and disrupts schedule
efficiency despite short travel distances.

\end{tcolorbox}
\caption{Case: Excessive crowd buffer inflating visit duration.}
\label{fig:case_example3}
\end{figure}

\begin{figure}[h]
\centering
\begin{tcolorbox}[enhanced,
    colback=yellow!15,
    colframe=black,
    boxrule=0.5pt,
    arc=1mm,
    width=0.45\textwidth,
    fontupper=\scriptsize]

\textbf{Query:} Plan a 3 days trip for one person from Vadodara
to Kanpur from 2025-11-18 to 2025-11-20, with a budget of
INR 32024.

\noindent\rule{\linewidth}{0.4pt}

\textbf{Plan:}

\texttt{\{}\\
\texttt{"days": 2,}\\
\texttt{"current\_city": "Kanpur",}\\
\texttt{"transportation": "-",}\\
\texttt{"breakfast": "Little Chef Restaurant, Kanpur",}\\
\texttt{"attraction": "Maharana Pratap Sagar, Kanpur",}\\
\texttt{"lunch": "Haveli Restaurant, Kanpur",}\\
\texttt{"dinner": "Dadi ki rasoi, Kanpur",}\\
\texttt{"accommodation": "The Landmark Towers, Kanpur",}\\
\texttt{"event": "-",}\\
\texttt{"point\_of\_interest": "The Landmark Towers,}\\
\texttt{\hspace{1em}stay from 08:00 to 09:30, nearest}\\
\texttt{\hspace{1em}transit: Naveen Market, 0.254km away;}\\
\texttt{\hspace{1em}Little Chef Restaurant, visit from}\\
\texttt{\hspace{1em}09:40 to 10:30, nearest transit:}\\
\texttt{\hspace{1em}Chunniganj, 0.496km away;}\\
\colorbox{yellow}{\texttt{\hspace{1em}Maharana Pratap Sagar, visit from}}\\
\colorbox{yellow}{\texttt{\hspace{1em}11:00 to 14:30 (includes 30 min buffer}}\\
\colorbox{yellow}{\texttt{\hspace{1em}for above average/high crowd in}}\\
\colorbox{yellow}{\texttt{\hspace{1em}evening hours), nearest transit: -,;}}\\
\colorbox{yellow}{\texttt{\hspace{1em}Haveli Restaurant, visit from 14:00}}\\
\colorbox{yellow}{\texttt{\hspace{1em}to 15:20, nearest transit: -,;}}\\
\texttt{\hspace{1em}Dadi ki rasoi, visit from 20:00 to}\\
\texttt{\hspace{1em}21:15, nearest transit: Kanpur}\\
\texttt{\hspace{1em}Anwarganj, 2.26km away; The Landmark}\\
\texttt{\hspace{1em}Towers, stay from 21:30 to 07:00,}\\
\texttt{\hspace{1em}nearest transit: Naveen Market,}\\
\texttt{\hspace{1em}0.254km away."}\\
\texttt{\}}

\noindent\rule{\linewidth}{0.4pt}

\textbf{Remark:} Temporal overlap between consecutive
activities --- the visit to Maharana Pratap Sagar is scheduled
from 11:00 to 14:30, while the subsequent lunch at Haveli
Restaurant begins at 14:00, creating a \textbf{30-minute
overlap}. This violates chronological feasibility, leaving no
travel time and implying the traveler must be at two locations
simultaneously, indicating an invalid schedule construction.

\end{tcolorbox}
\caption{Case: Temporal overlap between consecutive activities.}
\label{fig:case_example4}
\end{figure}

\clearpage
\twocolumn

\section{Annotator Details}
\label{app:annotator-details}
 All annotators were students affiliated with our research lab and voluntarily participated in the annotation process as part of their regular research activities. Therefore, no separate monetary compensation was provided specifically for the annotation task.
\subsection{Guidelines for Annotators}
 
The annotation process involves providing annotators
with a structured travel plan skeleton containing
pre-selected cities, accommodations, attractions, and
restaurants rather than a fully detailed itinerary.
Annotators are tasked with filling in realistic and
optimal timings for each activity based on their own
real travel experiences and practical knowledge. The
annotated plan must reflect feasible scheduling while
considering reference information and local preferences
such as cuisine type, attraction category, and traveler
personas (e.g., laidback, economical). Additionally,
common sense should be applied when assigning visit
durations and transit gaps, and any deviations from
suggested durations or costs must be justified. A
detailed breakdown of these annotation guidelines,
including priority handling, public transit
considerations, and documentation requirements, is
provided in Table~\ref{tab:annotation_guidelines}.

\begin{figure}[H]
\centering
\scriptsize
\includegraphics[width=\columnwidth]{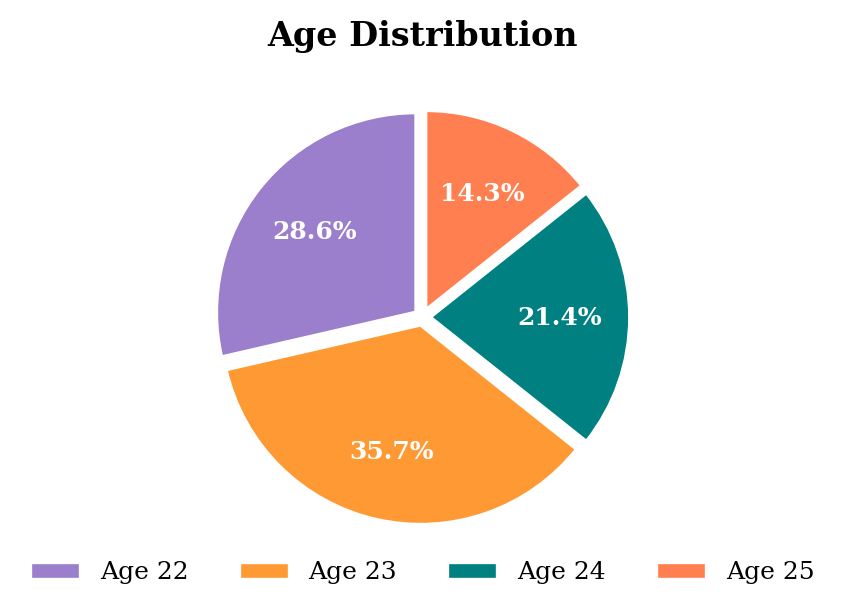}
\caption{Age distribution of our graduate student annotators.}
\label{fig:age_dist}
\end{figure}
 
\subsection{Annotator Demographics}
\label{app:annotator_demographics}
\begin{figure}
\centering
\includegraphics[width=\columnwidth]{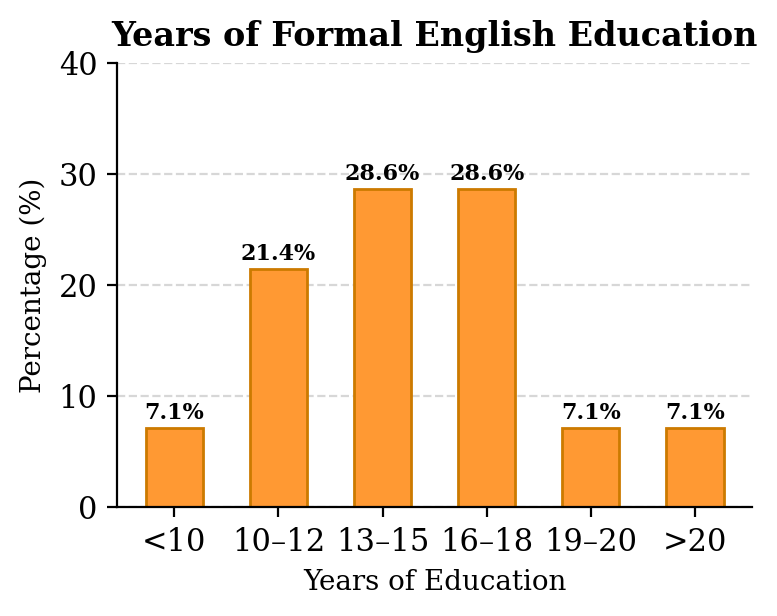}
\caption{Years of formal English education statistics of our
graduate student annotators.}
\label{fig:english_dist}
\end{figure}

\begin{figure}
\centering
\includegraphics[width=\columnwidth]{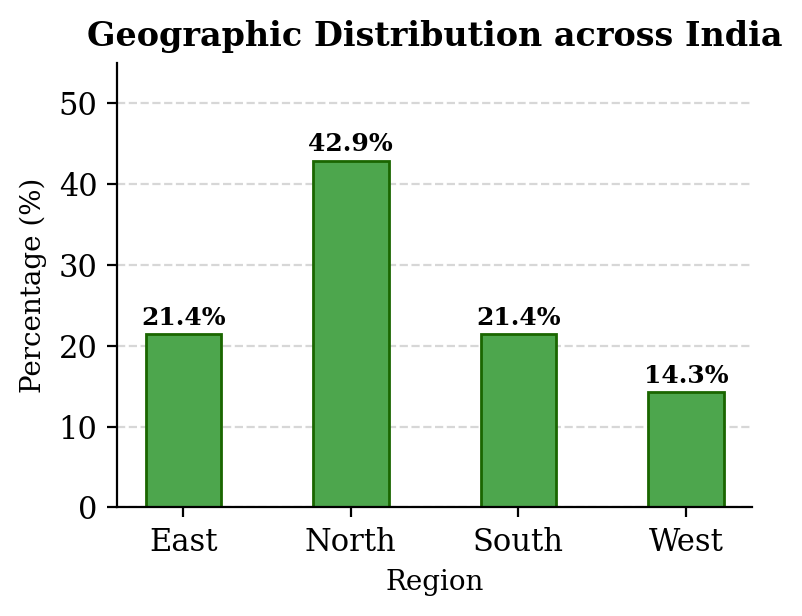}
\caption{Geographic distribution of annotators from diverse
regions of India.}
\label{fig:geo_dist}
\end{figure}

Total 14 trained human annotators participated in this study.
The annotator demographics, as illustrated by the
figures, show a diverse range of backgrounds. The age
distribution (Figure~\ref{fig:age_dist}) shows all
annotators fall between 22--25 years old, with the
majority aged 23 (35.7\%), followed by age 22
(28.6\%), age 24 (21.4\%), and age 25 (14.3\%),
indicating a young and energetic graduate student
cohort. The gender distribution
reflects balanced
participation, with 57.1\% male, 35.7\% female,
and 7.1\% other annotators. The years of English
education (Figure~\ref{fig:english_dist}) show that
most annotators have between 13--18 years of formal
English instruction, suggesting strong language
proficiency. We specifically recruited annotators from India only as \system{} is India-centric dataset. The geographic distribution
(Figure~\ref{fig:geo_dist}) also highlights that annotators
were recruited from diverse regions of India ---
North (42.9\%), East (21.4\%), South (21.4\%), and
West (14.3\%) --- ensuring a wide variety of
real-world travel experiences and regional familiarity,
contributing to reliable and contextually aware
annotations.

\begin{table*}[t]
\centering
\caption{Guidelines for Annotation of Travel Plans and Remarks.}
\label{tab:annotation_guidelines}
\begin{tabular}{p{0.05\linewidth} p{0.89\linewidth}}
\toprule
\textbf{\#} & \textbf{Annotation Guideline} \\
\midrule
\multicolumn{2}{l}{\textbf{Route \& Structure}} \\
1 & The trip must follow a closed-loop route: the first and last city must be the origin city (\texttt{org}). The city sequence must be logical with no unnecessary back-and-forth. \\
2 & Do not repeat restaurants, attractions, or events across the entire trip. Each restaurant, attraction, and event may appear only once in the full itinerary. \\
3 & The Point of Interest (PoI) list for each day must include \textit{all} activities: accommodation (check-in/stay), breakfast, lunch, dinner, attractions, and events, each with a valid timeline and transit info. \\
\midrule
\multicolumn{2}{l}{\textbf{Transportation \& Accommodation}} \\
4 & Transportation must be consistent. Day~1 must include valid transport from the origin. Do not mix conflicting modes (e.g., cab/taxi and flight). Mode must match \texttt{local\_constraint} if specified. \\
5 & For multi-night stays in the same city, use the same accommodation. Accommodations must be valid, verifiable places in the current city. \\
6 & At least 4 hours must separate the end of one meal from the start of the next (Breakfast $\rightarrow$ Lunch, Lunch $\rightarrow$ Dinner). The PoI timeline must reflect these gaps. \\
\midrule
\multicolumn{2}{l}{\textbf{Hard Constraints}} \\
7 & Hard constraints must be satisfied: total trip cost must not exceed the budget; cuisine, house rules, room type, facilities, and services specified in \texttt{local\_constraint} must all be respected. \\
\midrule
\multicolumn{2}{l}{\textbf{Uncertainty \& Buffer Times}} \\
8 & Use the provided Python scripts to compute mean buffer times for transportation and attractions/restaurants. These are reference values only; annotators may adjust based on trip context and traveler type. \\
9 & Buffer time should be scaled by traveler risk type --- \textit{Risk Tolerant}: $(0.5$--$1.0)\times$ ideal buffer; \textit{Risk Optimized}: $(1.0$--$1.25)\times$ ideal buffer; \textit{Risk Averse}: $(1.25$--$2.0)\times$ ideal buffer. \\
10 & Avoid scheduling attraction or restaurant visits during peak crowd hours where possible. Partial overlap with peak slots is acceptable if unavoidable (e.g., a 1:00--3:00\,PM visit when peak is 2:00--5:00\,PM is permitted). \\
\midrule
\multicolumn{2}{l}{\textbf{Documentation \& Flagging}} \\
11 &  Adjust the timings and Document the rationale for each timing choice in a Remarks column. \\
12 & If a repeated attraction is found across days, flag the row and note the repetition in the Remarks column (e.g., ``Attraction `X' repeated on Day~2 and Day~4''). \\
\bottomrule
\end{tabular}
\end{table*}

\end{document}